%% file: main.tex
\documentclass[10pt,twocolumn,letterpaper]{article}

\usepackage[pagenumbers]{cvpr} 

\usepackage{graphicx}
\usepackage{amsmath}
\usepackage{amssymb}
\usepackage{booktabs}
\input{math_commands.tex}

\usepackage{color}
\usepackage{xspace}
\makeatletter
\DeclareRobustCommand\onedot{\futurelet\@let@token\@onedot}
\def\@onedot{\ifx\@let@token.\else.\null\fi\xspace}
 
\def\ie{\emph{i.e}\onedot}

\def\etal{\emph{et al}\onedot}
\makeatother

\definecolor{darkorange}{rgb}{0.6, 0.15, 0.0}

\usepackage[pagebackref,breaklinks,colorlinks]{hyperref}

\usepackage[capitalize]{cleveref}
\crefname{section}{Sec.}{Secs.}
\Crefname{section}{Section}{Sections}
\Crefname{table}{Table}{Tables}
\crefname{table}{Tab.}{Tabs.}

\def\cvprPaperID{****} 
\def\confName{CVPR}
\def\confYear{2023}

\let\svthefootnote\thefootnote
\newcommand\blankfootnote[1]{%
  \let\thefootnote\relax\footnotetext{#1}%
  \let\thefootnote\svthefootnote%
}

\begin{document}

\title{MEGANE: Morphable Eyeglass and Avatar Network}


\author{Junxuan Li$^{1,2}$\thanks{Work done while Junxuan Li was an intern at Reality Labs Research, Pittsburgh, PA, USA}, Shunsuke Saito$^{2}$, Tomas Simon$^{2}$, Stephen Lombardi$^{2}$, Hongdong Li$^{1}$, Jason Saragih$^{2}$\\
$^1$Australian National University, \hspace{0.4cm} $^2$Meta Reality Labs Research
\vspace{0.1cm}\\
{ \url{junxuan-li.github.io/megane}}
}

\twocolumn[{%
\renewcommand\twocolumn[1][]{#1}%
\maketitle
\begin{center}
    \centering
\vspace{-1.0cm}
\includegraphics[width=0.97\textwidth]{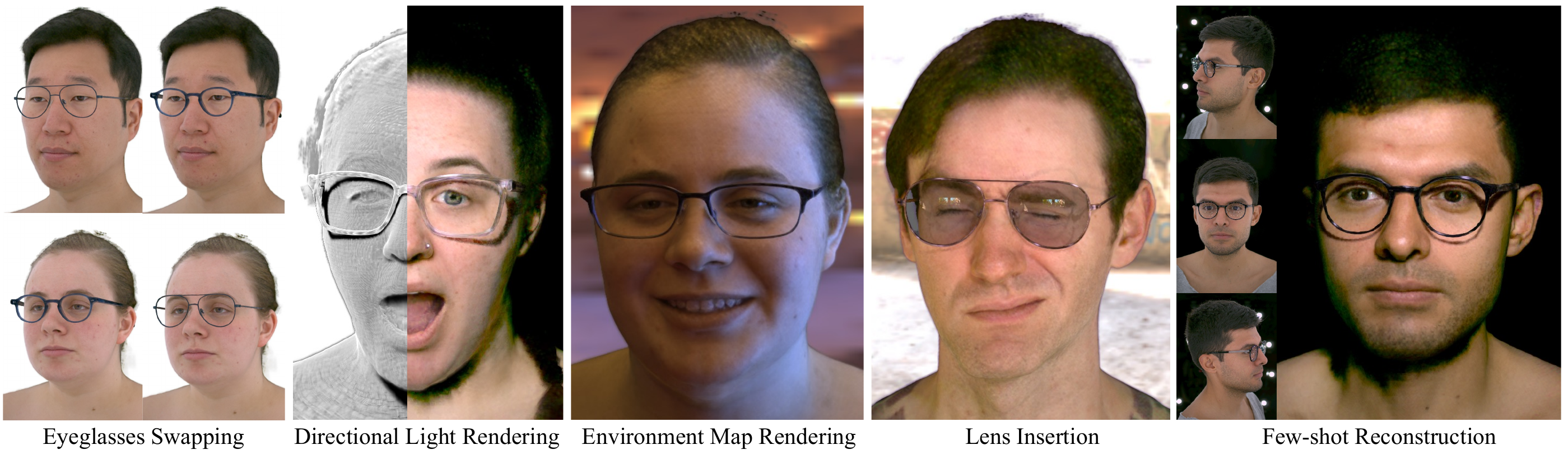}
\vspace{-0.3cm}
\captionof{figure}{\small{Our morphable eyeglasses supports the exchange of eyeglasses on face. Our relightable appearance correctly models glasses with different materials, and interactions between face and eyeglasses. In addition, our model enables lens insertion with appealing lens reflection and refraction effects. Once trained, our model can reconstruct and re-light an unseen eyeglasses with only a few inputs.}
}
\label{fig:teaser}
\end{center}%
}]

\begin{abstract}
\blankfootnote{* Work done while Junxuan Li was an intern at Reality Labs Research.}
Eyeglasses play an important role in the perception of identity. Authentic virtual representations of faces can benefit greatly from their inclusion.
However, modeling the geometric and appearance interactions of glasses and the face of virtual representations of humans is challenging. 
Glasses and faces affect each other's geometry at their contact points, and also induce appearance changes due to light transport.
Most existing approaches do not capture these physical interactions since they model eyeglasses and faces independently. Others attempt to resolve interactions as a 2D image synthesis problem and suffer from view and temporal inconsistencies.
In this work, we propose a 3D compositional morphable model of eyeglasses that accurately incorporates high-fidelity geometric and photometric interaction effects.
To support the large variation in eyeglass topology efficiently, we employ a hybrid representation that combines surface geometry and a volumetric representation.
Unlike volumetric approaches,
our model naturally retains correspondences across glasses, and hence explicit modification of geometry, such as lens insertion and frame deformation,  
is greatly simplified. 
In addition, our model is relightable under point lights and natural illumination, supporting high-fidelity rendering of various frame materials, including translucent plastic and metal within a single morphable model.
Importantly, our approach models global light transport effects, such as casting shadows between faces and glasses.
Our morphable model for eyeglasses can also be fit to novel glasses via inverse rendering.
We compare our approach to state-of-the-art methods and demonstrate significant quality improvements.

\end{abstract}

\section{Introduction}

Humans are social animals.
How we dress and accessorize is a key mode of self-expression and communication in daily life~\cite{flugel1933psychology}.
As social media and gaming has expanded social life into the online medium, virtual presentations of users have become increasingly focal to social presence, and with it, the demand for the digitization of clothes and accessories. 
In this paper, we focus on modeling eyeglasses, an everyday accessory for
billions
of people worldwide.

In particular, we argue that to achieve realism it is not sufficient to model eyeglasses in isolation: their interactions with the face have to be considered.
Geometrically, glasses and faces are not rigid, and they mutually deform one another at the contact points.
Thus, the shapes of eyeglasses and faces cannot be determined independently.
Similarly, their appearance is coupled via global light transport, and shadows as well as inter-reflections may appear and affect the radiance.
A computational approach to model these interactions is therefore necessary to achieve photorealism.

Photorealistic rendering of humans has been a focus of computer graphics for over 50 years, and yet the realism of avatars created by classical authoring tools still requires extensive manual refinement to cross the uncanny valley. 
Modern realtime graphics engines~\cite{metahuman} support the composition of individual components (e.g., hair, clothing), but the interaction between the face and other objects is by necessity approximated with overly simplified physically-inspired constraints or heuristics (e.g., ``no interpenetrations''). Thus, they do not faithfully reconstruct all geometric and photometric interactions present in the real world.

Another group of approaches aims to synthesize the composition of glasses in the image domain~\cite{lample2017fader,yao2021latent,xu2022temporally} by leveraging powerful 2D generative models~\cite{karras2020analyzing}. While these approaches can produce photorealistic images, animation results typically suffer from view and temporal inconsistencies due to the lack of 3D information.

Recently, neural rendering approaches~\cite{tewari2020neuralrendering} achieve photorealistic rendering of human heads~\cite{lombardi2018deep,lombardi2019neural,park2021nerfies,gafni2021dynamic,grassal2022neural} 
and general objects~\cite{niemeyer2020differentiable,yariv2020multiview,mildenhall2020nerf,wang2021neus} in a 3D consistent manner. 
These approaches are further extended to generative modeling for faces~\cite{cao2022authentic} and glasses~\cite{martin2020gelato,xie2021fig}, such that a single morphable model can span the shape and appearance variation of each object category. 
However, in these approaches~\cite{cao2022authentic,martin2020gelato,xie2021fig} interactions between objects are not considered, leading to implausible object compositions.
While a recent work shows that unsupervised learning of a 3D compositional generative model from an image collection is possible~\cite{niemeyer2021giraffe}, we observe that the lack of structural prior about faces or glasses leads to suboptimal fidelity.
In addition, the aforementioned approaches are not relightable, thus not allowing us to render glasses on faces in a novel illumination. 

In contrast to existing approaches, we aim at modeling the geometric and photometric interactions between eyeglasses frames and faces in a data-driven manner from image observations.
To this end, we present MEGANE (\textbf{M}orphable \textbf{E}ye\textbf{g}lass and \textbf{A}vatar \textbf{Ne}twork), a morphable and relightable eyeglass model that represents the shape and appearance of eyeglasses frames and its interaction with faces. 
To support variations in topology and rendering efficiency, we employ a hybrid representation combining surface geometry and a volumetric representation~\cite{lombardi2021mixture}.
As our hybrid representation offers explicit correspondences across glasses, we can trivially deform its structure based on head shapes.
Most importantly, our model is conditioned by a high-fidelity generative human head model~\cite{cao2022authentic}, 
allowing it to specialize deformation and appearance changes to the wearer. 
Similarly, we propose glasses-conditioned deformation and appearance networks for the morphable face model to incorporate the interaction effects caused by wearing glasses. 
We also propose an analytical lens model that produces photorealistic reflections and refractions for any prescription and simplifies the capture task, enabling lens insertion in a post-hoc manner.

To jointly render glasses and faces in novel illuminations, we incorporate physics-inspired neural relighting into our proposed generative modeling. 
The method infers output radiance given view, point-light positions, visibility, and specular reflection with multiple lobe sizes.
The proposed approach significantly improves generalization and supports subsurface scattering and reflections of various materials including translucent plastic and metal within a single model. 
Parametric BRDF representations can not handle such diverse materials, which exhibit significant transmissive effects, and inferring their parameters for photorealistic relighting remains challenging~\cite{zhang2021physg,zhang2021nerfactor,munkberg2022extracting}.

To evaluate our approach, we captured 25 subjects using a multi-view light-stage capture system similar to Bi~\etal~\cite{bi2021deep}. Each subject was captured three times; once without glasses, and another two times wearing a random selection out of a set of 43 glasses. All glasses were captured without lenses.
As a preprocess, we separately reconstruct glasses geometry using a differentiable neural SDF from multi-view images~\cite{wang2021neus}. 
Our study shows that carefully designed regularization terms based on this precomputed glasses geometry significantly improves the fidelity of the proposed model. 
We also compare our approach with state-of-the-art generative eyeglasses models, demonstrating the efficacy of our representation as well as the proposed joint modeling of interactions. 
We further show that our morphable model can be fit to novel glasses via inverse rendering and relight them in 
new illumination conditions.

In summary, the contributions of this work are:
\begin{itemize}
    \vspace{-2.5mm}
    \item the first work that tackles the joint modeling of geometric and photometric interactions of glasses and faces from dynamic multi-view image collections.
    \vspace{-2.5mm}
    \item a compositional generative model of eyeglasses that represents topology varying shape and complex appearance of eyeglasses using a 
    hybrid mesh-volumetric 
    representation.
    \vspace{-2.5mm}
    \item a physics-inspired neural relighting approach that supports global light transport effects of diverse materials in a single model.
    
\end{itemize}

\section{Related Work}
We discuss related work in facial avatar modeling, eyeglasses modeling, and image-based editing.

\smallskip \noindent \textbf{Facial Avatar Modeling.}
Modeling photorealistic human faces is a long standing problem in computer graphics and vision. Early works leverage multi-view capture systems to obtain high-fidelity human faces~\cite{Huang04,Zhang04,Pighin06,Bickel07,Furukawa09,Bradley10,Beeler11,Fyffe14}. While these approaches provide accurate facial reflectance and geometry, photorealistic rendering requires significant manual effort~\cite{Seymour17} and typically not real-time with physics-based rendering.
Later, the prerequisites of facial avatar modeling are reduced to monocular videos~\cite{ichim2015dynamic,garrido2013reconstructing,thies2016face2face,cao2016real}, RGB-D inputs~\cite{thies2018headon} or a single image~\cite{hu2017avatar,nagano2018pagan}. 
However, these approaches do not provides authentic reconstruction of avatars. 
Lombardi~\etal~\cite{lombardi2018deep} demonstrate photorealistic rendering of dynamic human faces in a data-driven manner using neural networks. The learning-based avatar modeling is later extended to volumetric representations~\cite{lombardi2019neural}, a mesh-volume hybrid representation~\cite{lombardi2021mixture}, and a tetrahedron-volume hybrid representation~\cite{garbin2022voltemorph}.
Bi~\etal~\cite{bi2021deep} enable high-fidelity relighting of photorealistic avatars in real-time. 
While the aforementioned approaches require multi-view capture systems, recent works show that modeling of photorealistic avatars from monocular video inputs is also possible~\cite{athar2021flame,gafni2021dynamic,grassal2022neural}.
Cao~\etal~\cite{cao2022authentic} recently extend these person-specific neural rendering approaches to a multi-identity model, and demonstrates the personalized adaptation of the learned universal morphable model from a mobile phone scan.
Notably, these learning-based photorealistic avatars neither study nor demonstrate the accurate composition of accessories including eyeglasses.

\smallskip \noindent \textbf{Eyeglasses Modeling.}
Eyeglasses are one of the most commonly used accessories in our daily life, and virtual try-on has been extensively studied~\cite{niswar2011virtual,huang2012human,zhang2017virtual,tang2014making,yuan2011mixed,huang2013vision,li2011eyeglasses}. 
An image-based eyeglasses try-on is possible by composing a glasses image onto a face using Poisson blending~\cite{li2011eyeglasses}.
3D-based solutions have been also proposed for virtual reality~\cite{niswar2011virtual} or mixed reality~\cite{yuan2011mixed} by leveraging predefined 3D eyeglasses models.
Zhang~\etal~\cite{zhang2017virtual} enable lens refraction and reflection in their proposed try-on system.
However, these approaches rely on predefined 3D glass models, and cannot represent novel glasses. In addition, supported frames are limited to non-transparent reflective materials and the fidelity is limited by real-time graphics engines.

Recent progress in neural rendering~\cite{mildenhall2020nerf,tewari2020neuralrendering,wang2021neus} enables photorealistic modeling of general 3D objects. 
Several works extend the neural rendering techniques to generative models to represent various shapes and materials of objects in the same category using a single model~\cite{martin2020gelato,xie2021fig}. GeLaTO~\cite{martin2020gelato} presents a billboard-based neural rendering method to represent different glasses. Fig-NeRF~\cite{xie2021fig} extends neural radiance fields (NeRF)~\cite{mildenhall2020nerf} to generative modeling. 
However, these methods individually model glasses and are not conditioned by the information of the wearers. Thus, the complex geometric and photometric interactions are not incorporated in the composition. 
More recent approaches learn to decompose multiple 3D objects in an unsupervised manner, allowing us to compose them with different combination~\cite{niemeyer2021giraffe,yang2021learning,wu2022object}.
GIRAFFE~\cite{niemeyer2021giraffe} models the scene as composition of multiple NeRFs using adversarial training. 
While these approaches are promising, we observe that lack of explicit structural prior leads to suboptimal decomposition, failing to model photorealistic interactions.

\smallskip \noindent \textbf{Generative Models.}
Generative models have demonstrated remarkable ability in synthesizing photorealistic images, including human faces\cite{karras2020analyzing}. 
Recent work has extended these models to add intuitive semantic editing, such as synthesis of glasses on faces~\cite{lample2017fader,he2019attgan,liu2019stgan,yao2021latent,xu2022temporally}.
Fader Networks~\cite{lample2017fader} disentangle the salient image information, and then generate different images by varying attribute values, including glasses on faces.
Subsequent work has proposed two decoders for modeling latent representations and facial attributes~\cite{he2019attgan}, selective transfer units~\cite{liu2019stgan}, and geometry-aware flow~\cite{yin2019instance} to further improve editing fidelity.
Yao~\etal~\cite{yao2021latent} extend facial attribute editing to video sequences via latent transformation and a identity preservation loss, which is further improved by Xu~\etal~\cite{xu2022temporally}, incorporating flow-based consistency.
More recent works propose 3D-aware generative models to achieve view-consistent synthesis~\cite{xiang2022gram,Deng_2022_CVPR,Chan_2022_CVPR,Or-El_2022_CVPR,epigraf,Xu_2022_CVPR,wang2022morf}. In particular, IDE-3D~\cite{sun2022ide} proposes a 3D-aware semantic manipulation. However, the precise modeling and relighting of interactions between glasses and faces has been neither studied nor demonstrated.

\smallskip\noindent\textbf{Image-based Relighting.}
Various image-based solutions have been proposed to enable human face relighting ~\cite{sun2019single,tewari2020stylerig,wang2020single,pandey2021total,yeh2022learning}.
Sun~\etal~\cite{sun2019single} enables image-based relighting using an encoder-decoder network. 
StyleRig~\cite{tewari2020stylerig} proposes a method to invert StyleGAN~\cite{karras2020analyzing} with explicit face prior, allowing the synthesizing pose or illumination changes for an input portrait.
Wang~\etal~\cite{wang2020single} and Total Relighting~\cite{pandey2021total} infer skin reflectances such as surface normal and albedo in the image space, and use them to generate shading and reflection, which are fed into network for better generalization.
Lumos~\cite{yeh2022learning} trains a relighting network on large-scale synthesized data and proposes several regularization terms to enable domain transfer to real portraits. 

While these image-based approaches successfully synthesize photorealistic interaction and relighting of glasses and faces, lack of 3D information including contact and occlusion leads to limited fidelity and incoherent results in motion and changing views.

\begin{figure*}
    \centering
    \includegraphics[width=1\textwidth]{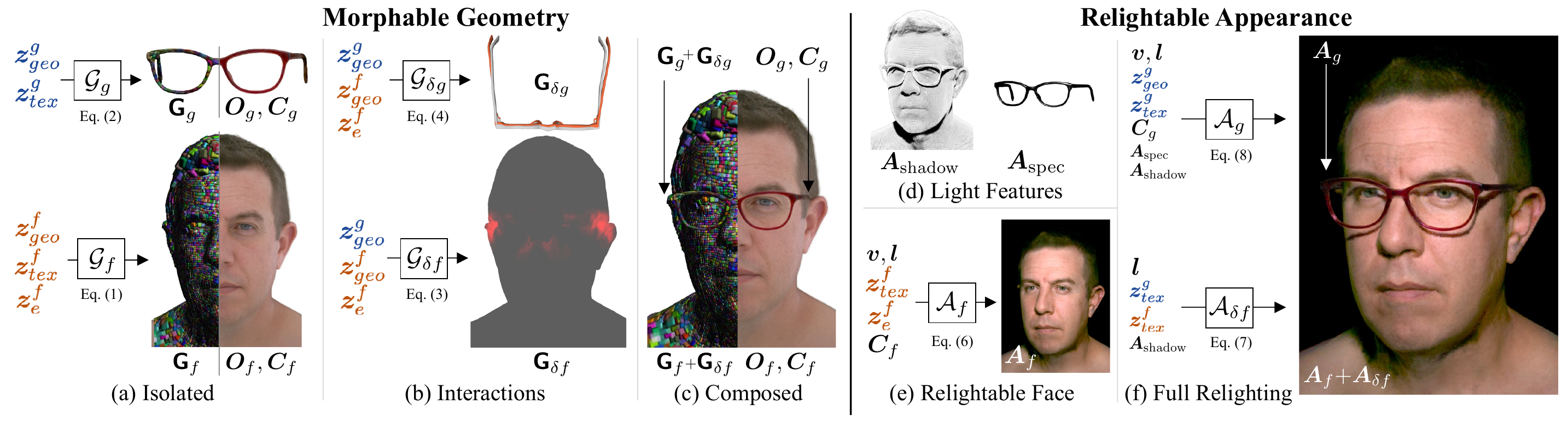}
    \vspace{-0.7cm}
    \caption{\textbf{Overview}. Our approach learns (a) separate latent spaces to model variations in faces and eyeglasses, as well as (b) their geometric interactions such that the models can be (c) composed together. Additionally, to accurately render relightable appearance, we compute features (d) that represent light interactions with (e) a relightable face model to allow for (f) joint face and eyeglass relighting.
    }
    \label{fig:overview}
    \vspace{-0.5cm}
\end{figure*}

\section{Method} 
Our method consists of two components, morphable geometry and relightable appearance, as shown in \cref{fig:overview}. 

\subsection{Morphable Geometry}
\label{sect:morphable_geometry}

Our approach is based on Mixture of Volumetric Primitives~(MVP)~\cite{lombardi2021mixture}, a distinct volumetric neural rendering approach that achieves high-fidelity renderings in real-time. Compared to neural fields approaches~\cite{xie2022neural}, it contains explicit volumetric primitives that move and deform to efficiently allow expressive animation with semantic correspondences across frames. Also unlike mesh-based approaches~\cite{lombardi2018deep}, it supports topological changes in geometry. 

To model faces without glasses, we adopt the pretrained face encoder $\gE_{f}$ and decoder $\gG_{f}$ from Cao~\etal~\cite{cao2022authentic}.
Given an encoding of the facial expression $\vz^f_e$, and face identity encoding of geometry $\vz^f_{geo}$ and textures $\vz^f_{tex}$, the face primitive geometry and appearance are decoded as:

\begin{align}
    \tG_{f}, \mO_{f}, \mC_{f} = \gG_{f} (\vz^f_e , \vz^f_{geo}, \vz^f_{tex}),
\end{align}
where $\tG_{f} = \{ \vt, \mR, \vs \}$ is the tuple of the position $\vt \in \sR^{3\times N_{\text{fprim}}}$, rotation $\mR \in \sR^{3\times 3 \times N_{\text{fprim}}}$ and scale $\vs \in \sR^{3\times N_{\text{fprim}}} $ of face primitives; $\mO_{f} \in \sR^{M^{3} \times N_{\text{fprim}}}$ is the opacity of face primitives; $\mC_{f} \in \sR^{3\times M^{3} \times N_{\text{fprim}}}$ is the RGB color of face primitives in fully-lit images. $N_{\text{fprim}}$ denotes the number of face primitives and $M$ denote the resolution of each primitives. We follow previous work~\cite{cao2022authentic} and use $N_{\text{fprim}}=128\times128$ and $M=8$. 

To model glasses, we propose a generative morphable eyeglass network that consists of a variational auto-encoder architecture:
$
\vz^g_{geo}, \vz^g_{tex} = \gE_g (\vw^g_{\text{id}}),
$
where $\gE_g$ is a glasses encoder that takes a one-hot-vector $\vw^g_{\text{id}}$ of glasses at input, and generates both geometry and appearance latent codes for the glasses $\vz^g_{geo}, \vz^g_{tex}$ as output. 
We then use the latent codes for a morphable glasses geometry decoder:
\begin{align}
    \tG_{g}, \mO_{g}, \mC_{g} = \gG_{g} (\vz^g_{geo}, \vz^g_{tex}),
\end{align}
where $\tG_{g} = \{ \vt_g, \mR_g, \vs_g \}$ is the tuple of the position, rotation and scale of the eyeglasses primitives, with position $\vt_g \in \sR^{3\times N_{\text{gprim}}}$, rotation $\mR_g \in \sR^{3\times 3 \times N_{\text{gprim}}}$ and scale $\vs_g \in \sR^{3\times N_{\text{gprim}}} $; $\mO_{g} \in \sR^{M^{3} \times N_{\text{gprim}}}$ the opacity of glasses primitives; $\mC_{\text{g}} \in \sR^{3\times M^{3} \times N_{\text{gprim}}}$ is the RGB color of glasses primitives in fully-lit images. $N_{\text{gprim}}$ denotes the number of glasses primitives; we use $N_{\text{gprim}}=32\times 32$.

We model the deformation caused by the interaction as residual deformation of the primitives:
\begin{align}
    \tG_{\delta{f}} &= \gG_{\delta{f}} (\vz^f_e , \vz^g_{geo}, \vz^f_{geo}),\\
    \tG_{\delta{g}} &= \gG_{\delta{g}} (\vz^f_e , \vz^g_{geo}, \vz^f_{geo}),
\end{align}
where $\tG_{\delta{f}} = \{ \delta_{\vt}, \delta_{\mR}, \delta_{\vs} \}, \tG_{\delta{g}} = \{ \delta_{\vt_g}, \delta_{\mR_g}, \delta_{\vs_g} \}$ are the residuals in position, rotation and scale from their values in the canonical space.
Specifically, the interaction influences the eyeglasses in two different ways: non-rigid deformations caused by fitting to the head, and rigid deformations caused by facial expressions. We found that individually modeling these two effects better generalize to a novel combination of glasses and an identity. Therefore, we model the deformation residuals as 
\begin{align}
    \gG_{\delta{g}} (\cdot) =  \gG_{\text{deform}} (\vz^g_{geo}, \vz^f_{geo}) +  \gG_{\text{transf}} (\vz^f_e , \vz^g_{geo})
\end{align}
where $\gG_{\text{deform}}$ takes facial identity information to deform the eyeglasses to the target head, and $\gG_{\text{transf}}$ takes expression encoding as input to model the relative rigid motion of eyeglasses on face caused by different facial expressions (e.g., sliding up when wrinkling the nose). 

\subsection{Relightable Appearance}
\label{sect:app}

The appearance model in previous works based on volumetric primitives~\cite{lombardi2021mixture,cao2022authentic} integrates the captured lighting environment as part of appearance, and cannot relight the avatar to novel illuminations. The appearance values of primitives under the uniform tracking illumination in Sec.~\ref{sect:morphable_geometry}, $\mC_{f}$ and $\mC_{g}$ are only used for learning geometry and the deformation by interactions. To enable relighting of the generative face model, we train a relightable appearance decoder that is additionally conditioned on view direction $\vv$ and light direction $\vl$ following~\cite{bi2021deep}:
\begin{align}
    \mA_{f} = \gA_{f} (\vz^f_e, \vv, \vl, \vz^f_{tex}, \mC_{f}),
\end{align}
where $\mA_{f} \in \sR^{3\times M^3 \times  N_{\text{fprim}}} $ is the appearance slab consists of RGB colors under a single point-light. 

To model the photometric interaction of eyeglasses on faces, we consider it as residuals conditioned by a eyeglasses latent code, similarly to the deformation residuals. 
Additionally, we observed that the most noticeable appearance interactions of eyeglasses on the face are from cast shadows. 
We explicitly provide shadow feature as an input to facilitate shadow modeling:
\begin{align}\label{eq:7}
    \mA_{\delta{f}} = \gA_{\delta{f}} (\vl, \vz^g_{tex},  \vz^f_{tex}, \mA_{\text{shadow}}) 
\end{align}
where 
$\mA_{\delta{f}} \in \sR^{3\times M^3 \times N_{\text{fprim}}}$ 
is the appearance residual for the face; and 
$\mA_{\text{shadow}}\in \sR^{M^3 \times  N_{\text{fprim}}} $ 
is the shadow feature computed by accumulating opacity while ray-marching from each of the light sources to the primitives, representing light visibility~\cite{lokovic2000deep}. Thus, the shadow feature represents the first bounce of light transport on both the face and glasses.

We model the relightalble glasses appearance similarly to the relightable face. Since this work focuses on modeling eyeglasses on faces, we define it as a conditional model with face so that occlusion and multiple bounces of lights by an avatar's head is already incorporated in the appearance:
\begin{align}
    \mA_{g} = \gA_{g} (\vv, \vl, \vz^g_{tex}, \vz^g_{geo}, \mA_{\text{shadow}}, \mA_{\text{spec}}, \mC_{g}).
\end{align}
where $\mA_{g} \in \sR^{3\times M^3 \times  N_{\text{gprim}}} $ is the glasses appearance slab, and $\mA_{\text{spec}} \in \sR^{3\times M^3 \times  N_{\text{gprim}}}$ is the specular feature; $\mA_{\text{shadow}}$ is the shadow feature computed in the same way as in \cref{eq:7}, which encodes face information.
We compute specular feature $\mA_{\text{spec}}$ at every point on primitives based on normal, light and view directions with a specular BRDF parameterized as Spherical Gaussians~\cite{li2022self} with three different lobes.
We observe that explicitly conditioning specular reflection significantly improves fidelity of relighting and generalization to various frame materials. Similar observations have been made for recent portrait relighting approaches~\cite{pandey2021total,yeh2022learning}.

\subsection{Differentiable Volumetric Rendering}
We render the predicted volumetric primitives following previous work~\cite{lombardi2021mixture}. Denote the position of all primitives in the space as $\tG$, when only render the face without wearing any eyeglasses, $\tG = \tG_{f}$;  and when wearing glasses $\tG = \{ \tG_{f}+\tG_{\delta{f}}, \tG_{g}+\tG_{\delta{g}} \}$. Denote the opacity of all primitives as $\mO$, it takes form $\mO=\mO_{f}$ or $\mO=\{ \mO_{f}, \mO_{g}\}$ for without and with glasses. Denote the color of all primitives as $\mC$, $\mC = \mC_{f}$ and $\mC = \{ \mC_{f}, \mC_{g}\}$ in fully-lit images, while $\mC = \mA_{f}$ and $\mC = \{ \mA_{f}+\mA_{\delta{f}}, \mA_{g}\}$ in relighting frames. 
We then use volumetric aggregation~\cite{lombardi2021mixture} to render images.

\subsection{Data Acquisition}
\label{sect:data}
\begin{figure}
    \centering
    \includegraphics[width=0.48\textwidth]{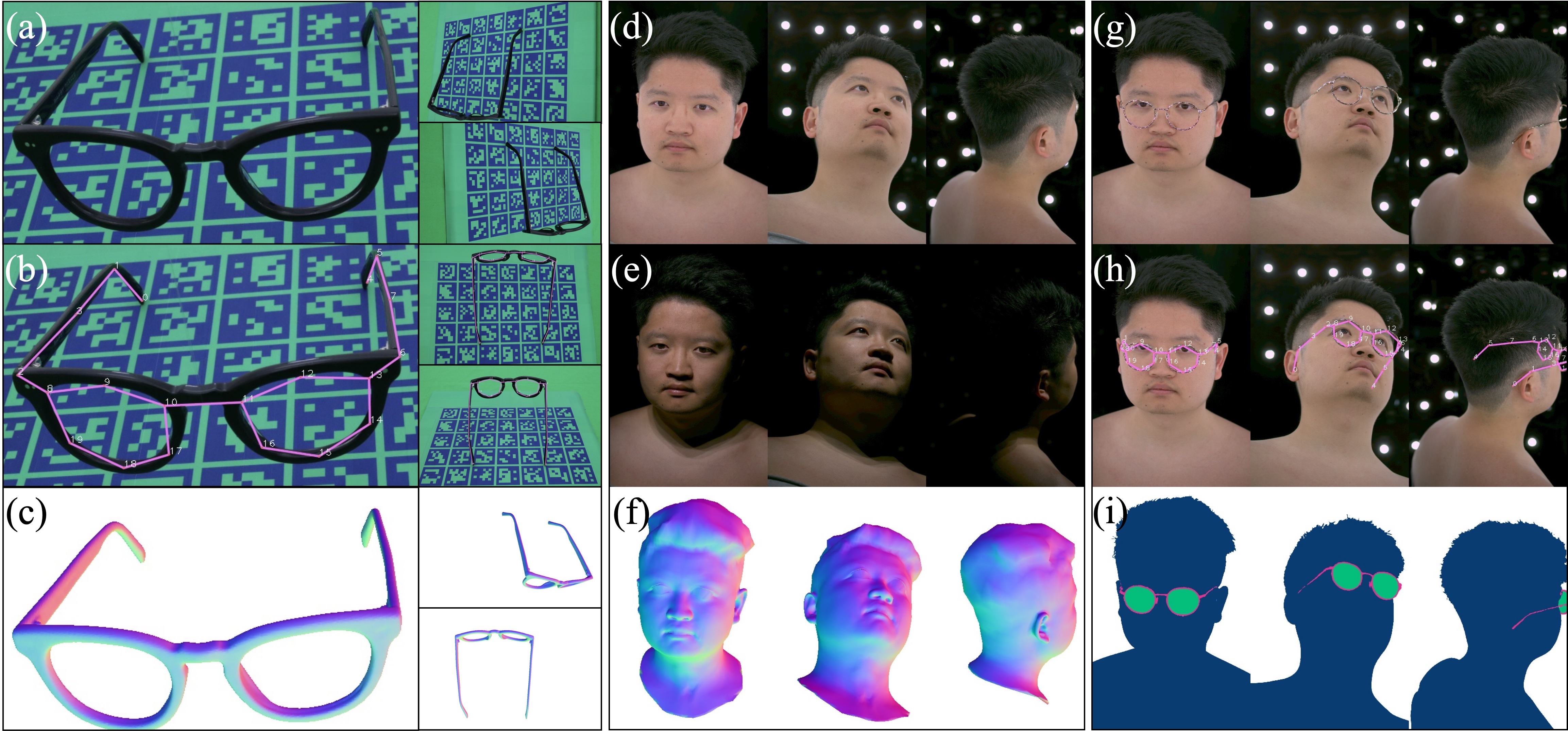}
    \caption{Datasets for {\em Eyeglasses} (a-c), {\em Faces} (d-f), and {\em Faces with Eyeglasses} (g-i). See text for description.}
    \label{fig:datasets}
    \vspace{-0.5cm}
\end{figure}
We aim to learn a generative model of eyeglasses and faces as well as the interactions between them. Therefore, we capture three types of data: {\em Eyeglasses}, {\em Faces}, and {\em Faces with Eyeglasses}. 
To decouple learning frame style from lens effects (which vary across prescriptions), we remove the lenses from the eyeglasses for all datasets.

\noindent \textbf{Eyeglasses}. We selected a set of 43 eyeglasses to cover a wide range of sizes, styles, and materials, including metal and translucent plastics of various colors. 
For each eyeglasses instance, we capture approximately 70 multi-view images using a hand-held DSLR camera (Fig.~\ref{fig:datasets}(a)). 
We apply a surface reconstruction method~\cite{wang2021neus} to extract 3D meshes of the eyeglasses (Fig.~\ref{fig:datasets}(c)). These 3D meshes will later provide supervision for the eyeglasses MVP geometry. However, because the glasses will change geometrically once they are worn, we use Bounded Biharmonic Weights (BBW)~\cite{jacobson2011bounded} to define a coarse deformation model that will be used to fit these meshes to the {\em Face With Eyeglasses} dataset using keypoint detections (Fig.~\ref{fig:datasets}(b)). 
Please see \cref{sec:DataAcquisition} for details of eyeglasses mesh reconstruction and registration.

\noindent \textbf{Faces} and \textbf{Faces with Eyeglasses}. 
We capture a dataset of faces without eyeglasses and the same set of faces with eyeglasses. This dataset consists of 25 subjects captured using a multi-view light-stage capture system with 110 cameras. Participants are instructed to perform various facial expressions, yielding recordings with changing expressions and head pose ((Fig.~\ref{fig:datasets}(d)).
Each subject was captured three times: once without glasses, and another two times wearing a random selection out of the set of 43 glasses (Fig.~\ref{fig:datasets}(g)). 

To allow for relighting, this data is captured under different illumination conditions. Similar to Bi~\etal~\cite{bi2021deep}, the capture system uses time-multiplexed illuminations. In particular, fully-lit frames, \ie frames for which all lights on the lightstage are turned on, are interleaved every third frame to allow for tracking, and the remaining two thirds of the frames are used to observe the subject under changing lighting conditions where only a subset of lights (``group'' lights) are turned on~(Fig.~\ref{fig:datasets}(e)).

Similar to prior work~\cite{lombardi2021mixture,cao2022authentic}, we first pre-process the data using a multiview face tracker to generate a coarse but topologically consistent face mesh for each frame (Fig.~\ref{fig:datasets}(f)). Tracking and detections are performed on fully lit frames and interpolated to partially lit frames when necessary.
Additionally, for the {\em Faces with Eyeglasses} portion, we detect a set of 20 keypoints on the eyeglasses~\cite{Li2019} (Fig.~\ref{fig:datasets}(h)) as well as face and glasses segmentation masks~\cite{Kirillov2019} (Fig.~\ref{fig:datasets}(i)), which are used to fit the eyeglasses BBW mesh deformation model to match the observed glasses.

\subsection{Training and Losses}
\label{sect:train}
We train the networks in two stages. In the first stage we use the fully-lit images to train the geometry of faces and glasses. Then, we use the images under group lights to train the relightable appearance model.

\noindent
{\bf Morphable Geometry Training}
We denote the parameters of the expression encoder in $\gE_{f}$, glasses encoder $\gE_g$, and decoders $\gG_{f}, \gG_{g}, \gG_{\delta{f}}, \gG_{\delta{g}}$ as $\Phi_g$, and optimize them using:
\begin{align}
    \Phi_g' = \argmin_{\Phi_g} \sum_{ N_I} \sum_{ N_{F_i}} \sum_{ N_C} \gL_{\text{fully-lit}}(\Phi_g, \mI^{i,r}), 
\end{align}
over $N_I$ different subjects; $N_{F_i}$ different fully-lit frames including with and without glasses; and $N_C$ different camera view points; and $\mI^{i}$ denotes all the ground truth camera images and associated processed assets for a frame, including face geometry, glasses geometry, face segmentation, and glasses segmentation; likewise, $\mI^{r}$ denotes the reconstructed images from volumetric rendering and the corresponding assets. 
Our fully-lit loss function consists of three main components:
\begin{align}
    \gL_{\text{fully-lit}}(\cdot) {=} \gL_{\text{rec}}(\mI^{i,r}) {+} \gL_{\text{gls}}(\mI^{i,r}) {+} \gL_{\text{reg}}(\Phi_g,\mI^{i,r}),
\end{align}
where the $\gL_{\text{rec}}$ are photometric reconstruction losses:
\begin{align}
    \gL_{\text{rec}}(\cdot) = \gL_{\text{L1}}(\mI^{i,r}) + \gL_{\text{vgg}}(\mI^{i,r}) + \gL_{\text{gan}}(\mI^{i,r}),
\end{align}
where $\gL_{\text{L1}}$ is the $l_1$ loss between observed images and reconstruction; $\gL_{\text{vgg}},\gL_{\text{gan}}$ are the VGG and GAN loss in~\cite{cao2022authentic}.

We also propose a geometry guidance loss $\gL_{\text{gls}}$ using the separately reconstructed glasses (Sec.~\ref{sect:data}) to improve the geometric accuracy of glasses, leading to better separations of faces and glasses in the joint training:
\begin{align}
    \gL_{\text{gls}}(\cdot) = 
    \gL_{\text{c}}(\mI^{i,r}) +
    \gL_{\text{m}}(\mI^{i,r}) + 
    \gL_{\text{s}}(\mI^{i,r})
\end{align}
including chamfer distance loss $\gL_{\text{c}}$; 
glasses masking loss $\gL_{\text{m}}$;
and glasses segmentation loss $\gL_{\text{k}}$. These losses encourage the network to separate identity-dependent deformations from glasses intrinsic deformations, thus helping the networks to generalize on different identities.
Please see \cref{sec:trainingandlosses} for details. 

In addition, we propose a regularization loss $\gL_{\text{reg}}$ for training: we use $\gL_{\text{KL}}(\cdot)$
the KL-divergence loss between the prior Gaussian distribution and the distribution of the glasses latent space; we also use a $l_2\text{-norm}$ for suppressing the delta deformation of face to reduce large displacements of face primitives. 
\begin{align}
    \gL_{\text{reg}}(\cdot) = 
    \gL_{\text{KL}} (\Phi_g) + 
    \gL_{\text{L2}} (\Phi_g, \mI^{i,r}).
\end{align}

During training, we set the weights of each loss term as 
$\lambda_{\text{L1}}=1,\lambda_{\text{vgg}}=1,\lambda_{\text{gan}}=1,\lambda_{\text{c}}=0.01,\lambda_{\text{m}}=10,\lambda_{\text{s}}=10,\lambda_{\text{KL}}=10^{-4},\lambda_{\text{L2}}=10^{-3}$. 
We train the first stage on a Nvidia Tesla V100 GPU with a batch size of 4 for 300k iterations using Adam optimizer~\cite{kingma2014adam} with a learning rate of $10^{-3}$, which takes around four days.

\noindent
{\bf Relightable Appearance Training}
Once the geometry module is trained, we freeze the parameters $\Phi_g$ and start training the relightable appearance $\gA_{f}$, $\gA_{\delta{f}}$, and $\gA_{g}$. We denote their parameters as $\Phi_a$,
We optimize the parameters $\Phi_a$ as follows:
\begin{align}
    \Phi_a' = \argmin_{\Phi_a} \sum_{ N_I} \sum_{ N_{G_i}} \sum_{ N_C} \gL_{\text{group-lit}}(\Phi_a, \mI^{i,c}), 
\end{align}
over $N_I$ different subjects; $N_C$ different cameras; and $N_{G_i}$ different group-light frames including with and without wearing glasses on face.

For frames illuminated by group-lights, we take the two nearest fully-lit frames to generate face and glasses geometry using  $\gG_{f}, \gG_{g}, \gG_{\delta{f}}, \gG_{\delta{g}}$, and linearly interpolate to get face and glasses geometry for the group-light image. 

The objective function for the second stage is mean-square-error photometric loss
$\gL_{\text{group-lit}}(\cdot) = || I^i - I^c ||_2^2$. The VGG and GAN loss are not used in relightable appearance training since we observe that these loss introduced block-like artifacts in the reconstruction. We use the same optimizer and GPU as in the previous stage. 
We train the second stage with a batch size of 3 for 200k iterations, which takes around four days.

\section{Experiments}
In this section, we evaluate each component of our method using the dataset of {\em Faces with Eyeglasses} and compare extensively with SOTA approaches. We exclude a set of frames and cameras for evaluation.

\begin{table}[]
\centering
\small{
\begin{tabular}{@{}c|cccc@{}}
\toprule
Components  & $l_1$($\downarrow$) & PSNR($\uparrow$) & SSIM($\uparrow$) & LPIPS($\downarrow$) \\ \midrule
w/o Geo     & 2.374              & 32.55              & 0.8764                & 0.1712            \\
w/o $\mA_{\text{shadow}}$    & 1.870          & 36.63          & 0.9227          & 0.1171          \\
w/o $\mA_{\text{spec}}$    & 1.577          & 37.69          & 0.9377          & 0.1087          \\
Full method & \textbf{1.558} & \textbf{37.98} & \textbf{0.9388} & \textbf{0.1034} \\ \bottomrule
\end{tabular}
}
\caption{Quantitative ablation of each part of our model.}
\label{tab:ablation}
\vspace{-0.5cm}

\end{table}

\subsection{Ablation Study}
\begin{figure}
    \centering
    \includegraphics[width=0.48\textwidth]{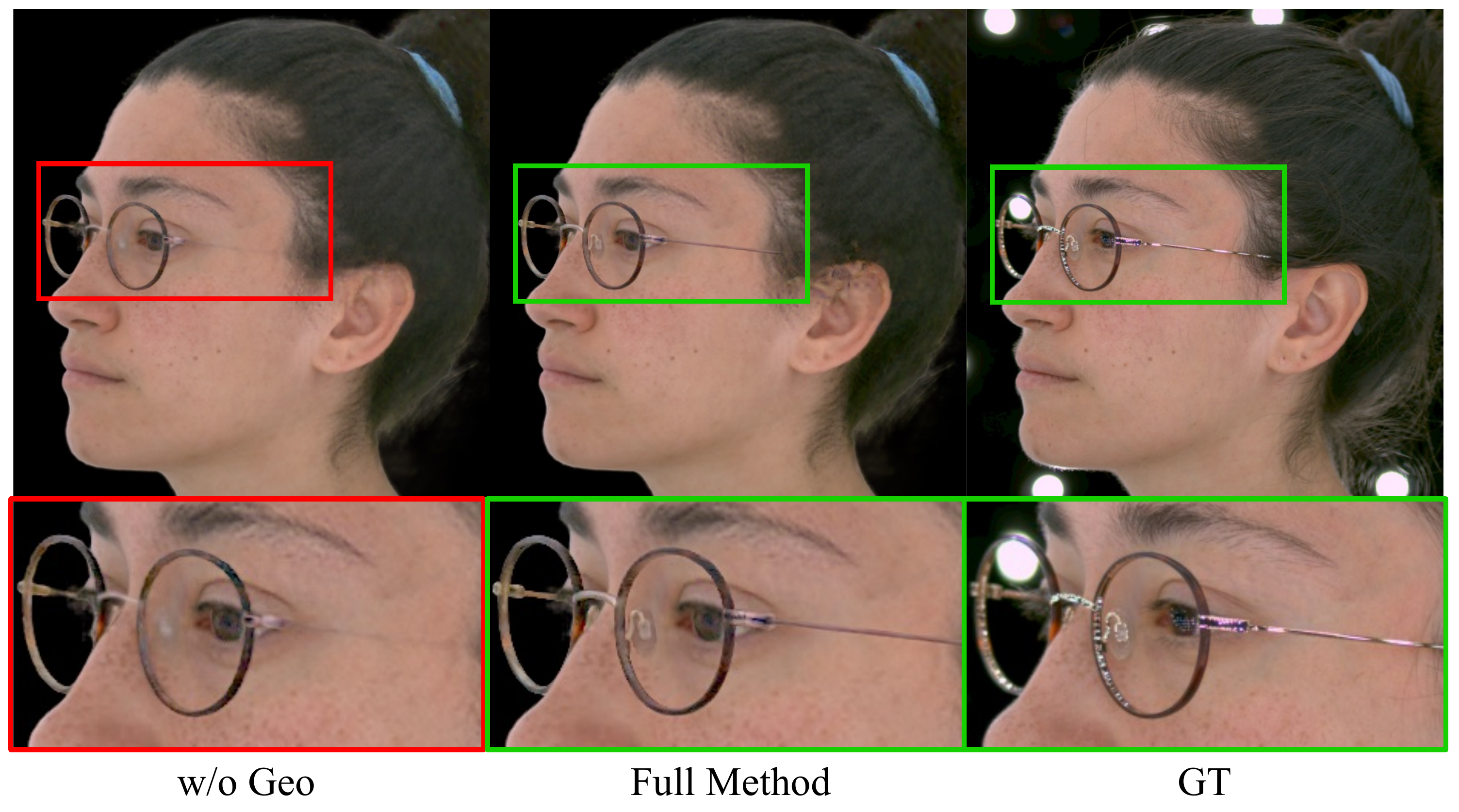}
    \vspace{-0.3cm}
    \caption{\textbf{Ablation study on geometry guidance.} Without geometry guidance lead to blurry results while our full model generates sharp and accurate eyeglasses.
    }
    \label{fig:nogeo_geo}
    \vspace{-0.3cm}
\end{figure}

\noindent
\textbf{Geometry Guidance.}
We first show that the proposed geometry-guided losses, including surface normal and segmentation, is essential for achieving crisp and sharp eyeglasses reconstruction. 
As shown in \cref{fig:nogeo_geo} and \cref{tab:ablation}, the model without using geometry guidance is only trained with image-based reconstruction and regularization losses. And it fails to reconstruct the detailed geometry of the eyeglasses, such as the nose pads. In comparison, the model with geometry guidance achieves higher geometric fidelity.

\begin{figure}
    \centering
    \includegraphics[width=0.48\textwidth]{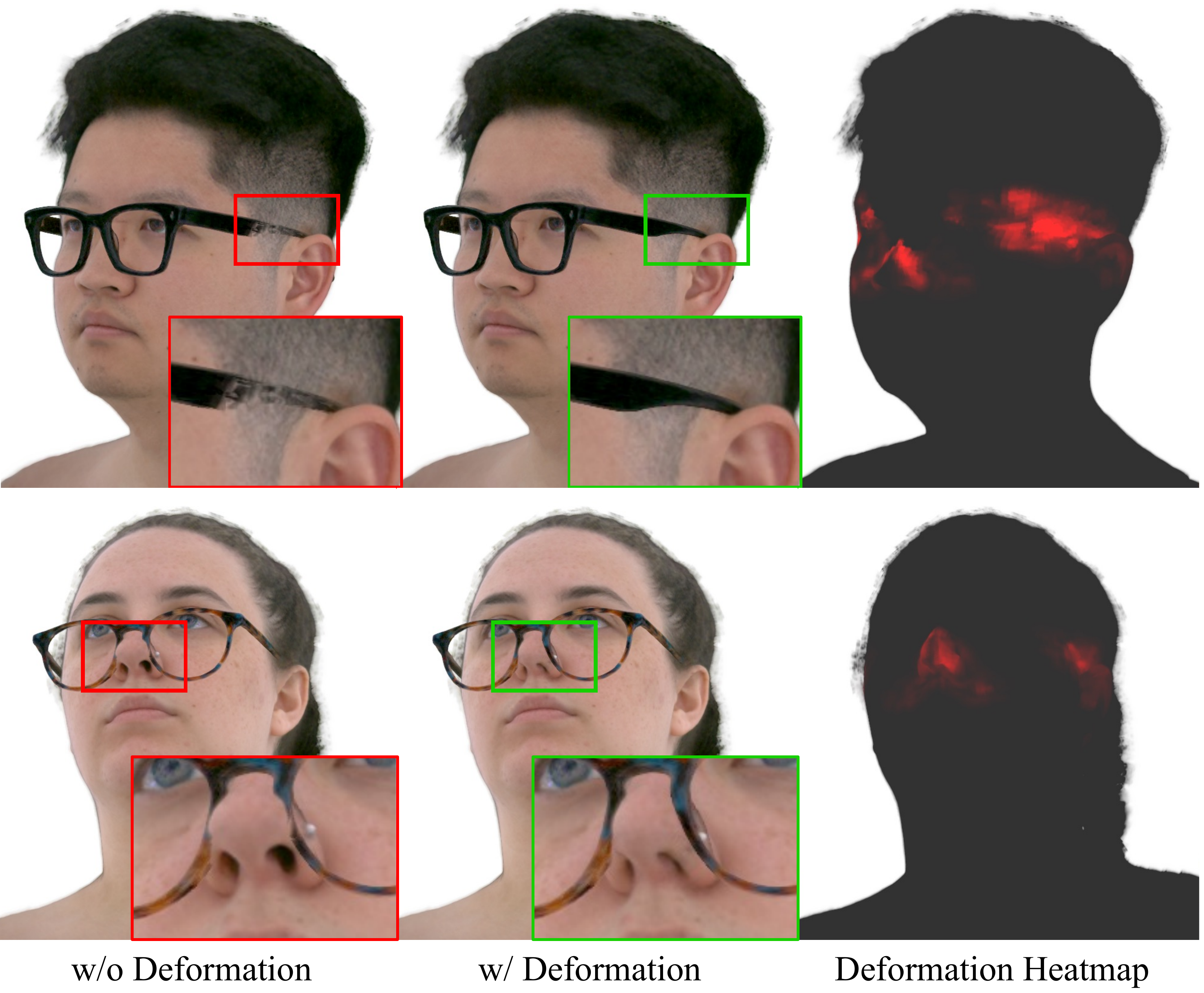}
    \caption{\textbf{Effectiveness of deformation. } 
    Face deformation modeling is critical for correctly rendering eyeglasses and face. 
    }
    \label{fig:nodeform}
    \vspace{-0.5cm}
\end{figure}

\noindent
\textbf{Geometry Interaction.} 
Eyeglasses and faces deform each other at contact points. We show in \cref{fig:nodeform} that without modeling such deformations, the legs of eyeglasses are rendered incorrectly and penetrate into the head. 
With the modeling of geometric interactions, our method learns and faithfully represents the deformation of the head as well as the nose.

\begin{figure}
    \centering
    \includegraphics[width=0.48\textwidth]{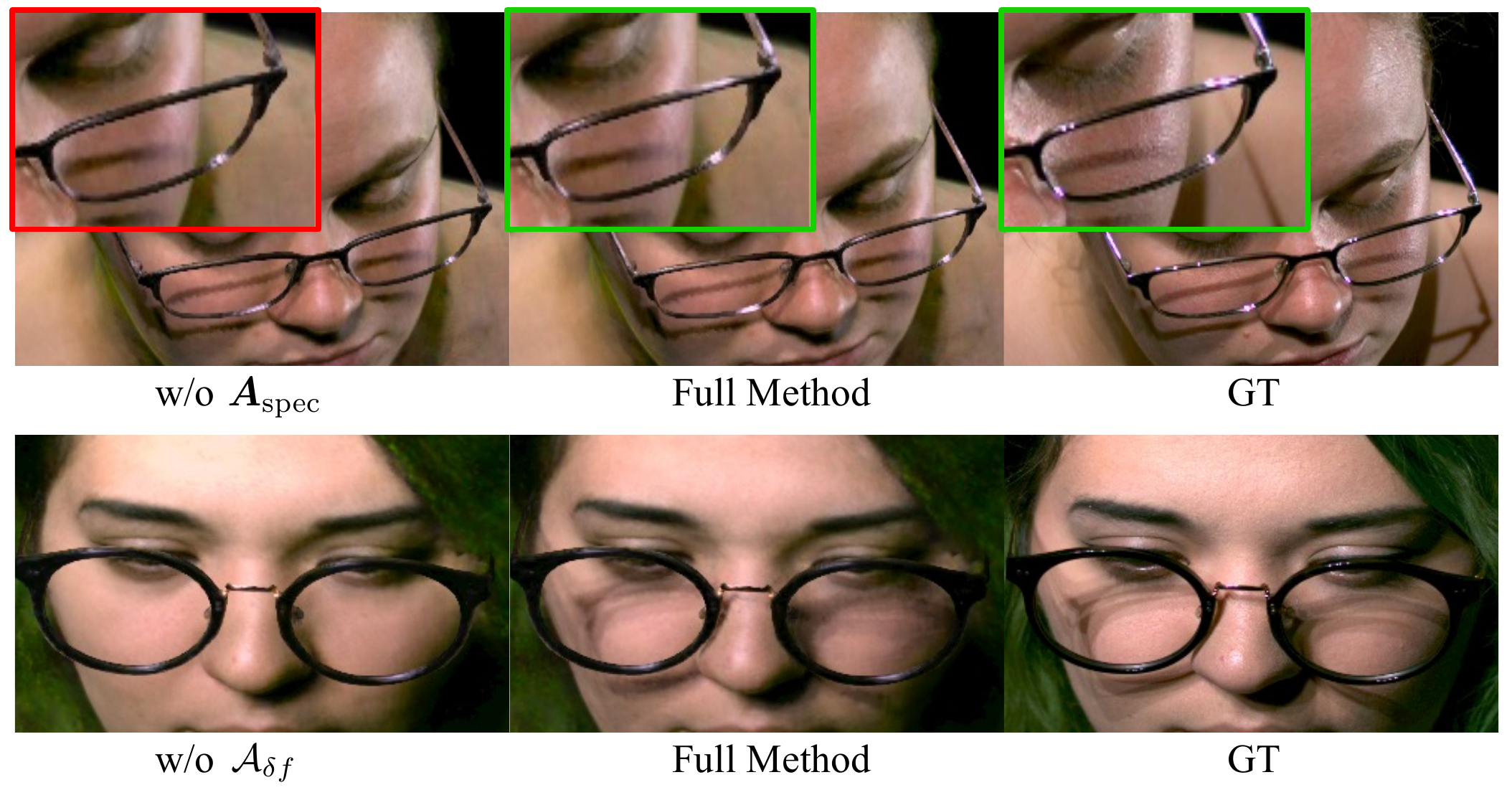}
    \caption{\textbf{Ablation study on specular feature and appearance interaction. }
    Top row: w/o using specular feature and full model. Bottom row: w/o appearance interaction and full model. 
    }
    \label{fig:noshad_nospec}
    \vspace{-0.1cm}
\end{figure}

\noindent
\textbf{Physics-inspired features for neural relighting.}
Here, we evaluate the effectiveness of the proposed specular and shadow feature on neural relighting. As shown in \cref{fig:noshad_nospec},
the one without using specular features fails to reconstruct specular highlights on the frame. Furthermore, the model without appearance interaction fails to reconstruct correct shadows on the face. 
We test and evaluate these components on held-out frames and present the quantitative results on \Cref{tab:ablation}. 
Adding each component effectively improves the performance on all metrics.

\subsection{Comparison}

\begin{figure}
    \centering
    \includegraphics[width=0.48\textwidth]{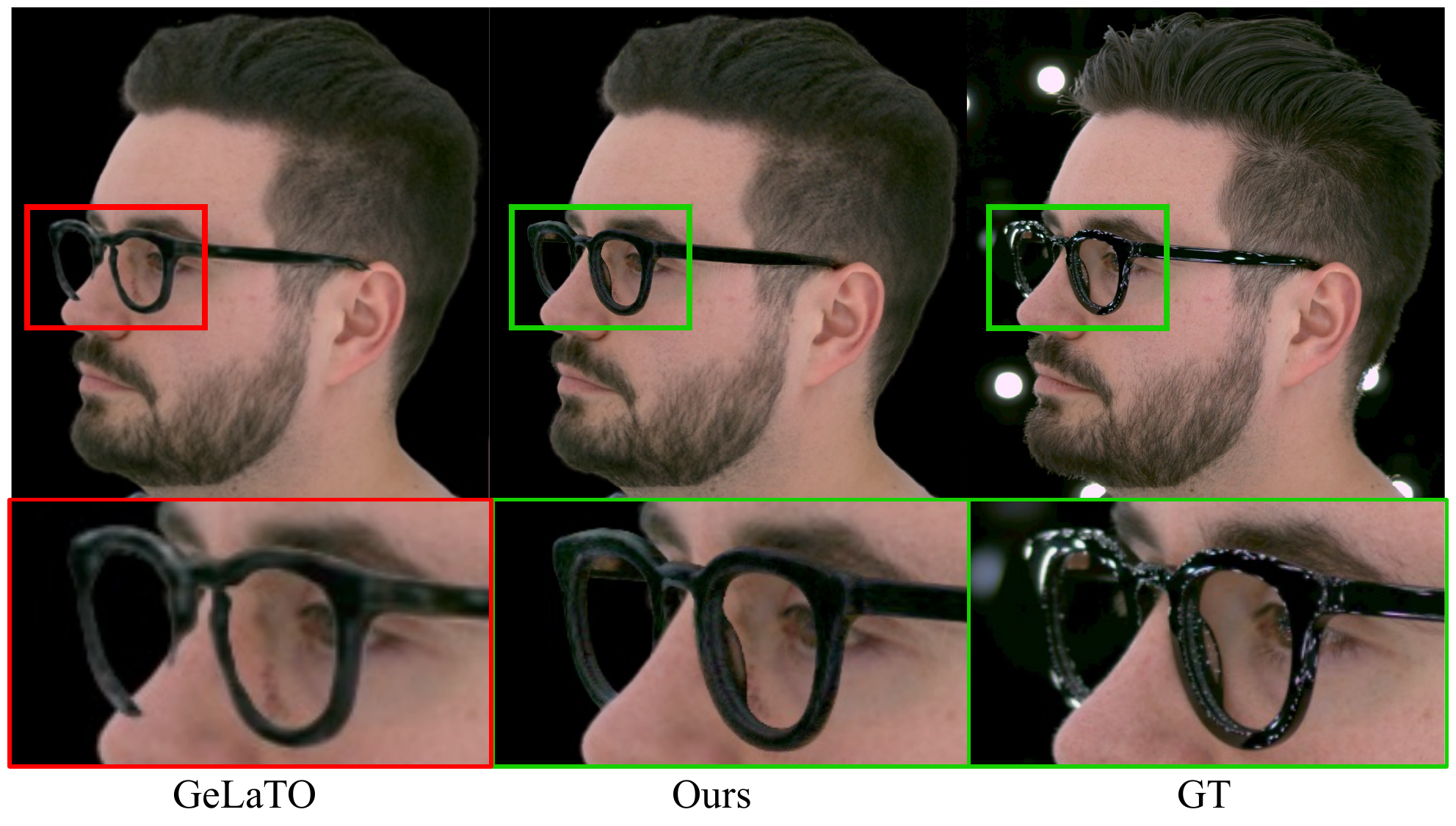}
    \caption{\textbf{Comparison with GeLaTO}~\cite{martin2020gelato}. Due to the simplified geometry representation, GeLaTO lacks geometry details and suffers from inconsistent occlusions.
    }
    \label{fig:gelato}
    \vspace{-0.1cm}
\end{figure}

\noindent
\textbf{GeLaTO~\cite{martin2020gelato}.} 
Previous work~\cite{martin2020gelato,xie2021fig} enables generative modeling of eyeglasses, but assume that everything except the glasses are static in the scene. In particular, Fig-NeRF~\cite{xie2021fig} is not applicable to our setup with severe occlusions and head motion. 
For comparison, we reimplement GeLaTO~\cite{martin2020gelato} and train with our datasets. Since GeLaTO does not support relighting, we compare only on fully-lit frames. \cref{fig:gelato} shows that while GeLaTO lacks geometric details and generates incorrect occlusion boundaries due to the billboard-based geometry, our method achieves high-fidelity results and correctly handles occlusions. \cref{tab:gelato} shows that our method also outperforms in all metrics.

\begin{table}[]
\centering
\small{
\begin{tabular}{@{}c|cccc@{}}
\toprule
Methods  & $l_1$($\downarrow$) & PSNR($\uparrow$) & SSIM($\uparrow$) & LPIPS($\downarrow$) \\ \midrule
GeLaTO~\cite{martin2020gelato}  & 16.561                 & 18.91                    & 0.6479                   & 0.2576                    \\
Ours    & \textbf{9.202}         & \textbf{21.80}           & \textbf{0.7690}          & \textbf{0.1614}           \\ \bottomrule
\end{tabular}
}
\caption{Quantitative comparison with GeLaTO.}
\label{tab:gelato}
\vspace{-0.1cm}
\end{table}

\begin{figure}
    \centering
    \includegraphics[width=0.48\textwidth]{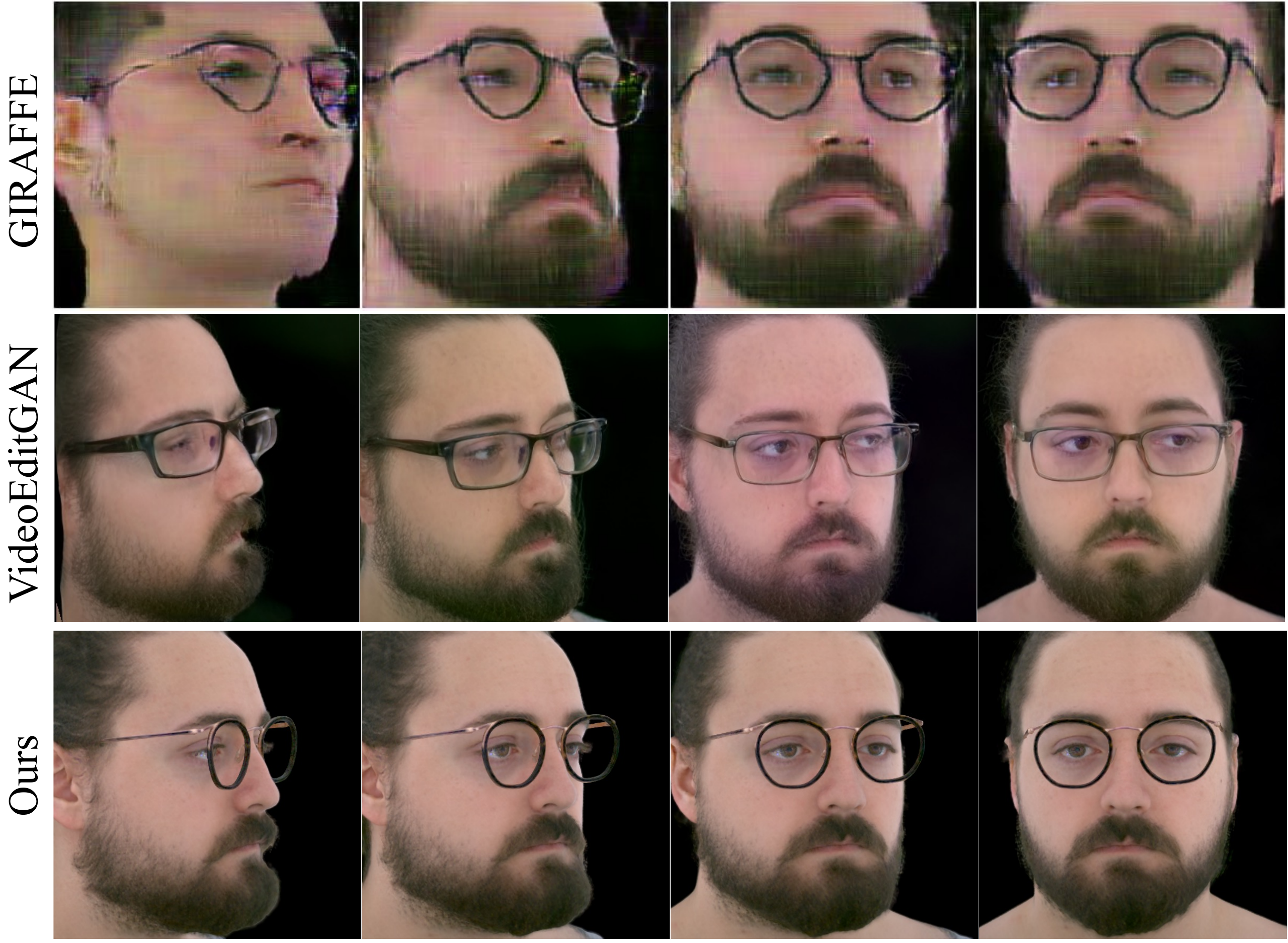}
    \caption{\textbf{Comparison with GIRAFFE~\cite{niemeyer2021giraffe} and VideoEditGAN~\cite{xu2022temporally}. }
    Compared with our method, other methods fail to render view consistent results.
    }
    \label{fig:videogan}
    \vspace{-0.1cm}
\end{figure}

\noindent
\textbf{GIRAFFE~\cite{niemeyer2021giraffe}} proposed a compositional neural radiance field that supports adding and changing objects in a scene. However, the official implementation only supports objects within the same category. For a fair comparison, we adapt their method to support adding generative objects in multiple categories. \cref{fig:videogan} shows that compositional generative modeling in an unsupervised manner still leads to suboptimal fidelity with limited resolution.

\noindent
\textbf{VideoEditGAN~\cite{xu2022temporally}} is a SOTA image-based editing method that allows us to insert glasses on face images. As shown in \cref{fig:videogan}, the image-based approach fails to maintain color and view consistency. Moreover, the approach cannot choose a specific type of glasses.
In contrast, our proposed representation enables the accurate reproduction of glasses and faces with consistent rendering in both view and time.

\noindent
\textbf{Relighting Comparison with Lumos~\cite{yeh2022learning}}. All the methods mentioned above do not support relighting of faces and eyeglasses. 
We compare our relighting results with Lumos~\cite{yeh2022learning}, a SOTA approach for portrait relighting. 
Due to the lack of 3D information, Lumos has difficulty rendering non-local light transport effects such as shadows cast by eyeglasses. 
In contrast, our method generates plausible soft shadows and accurately models photometric interactions between faces and glasses.

\begin{figure}
    \centering
    \includegraphics[width=0.48\textwidth]{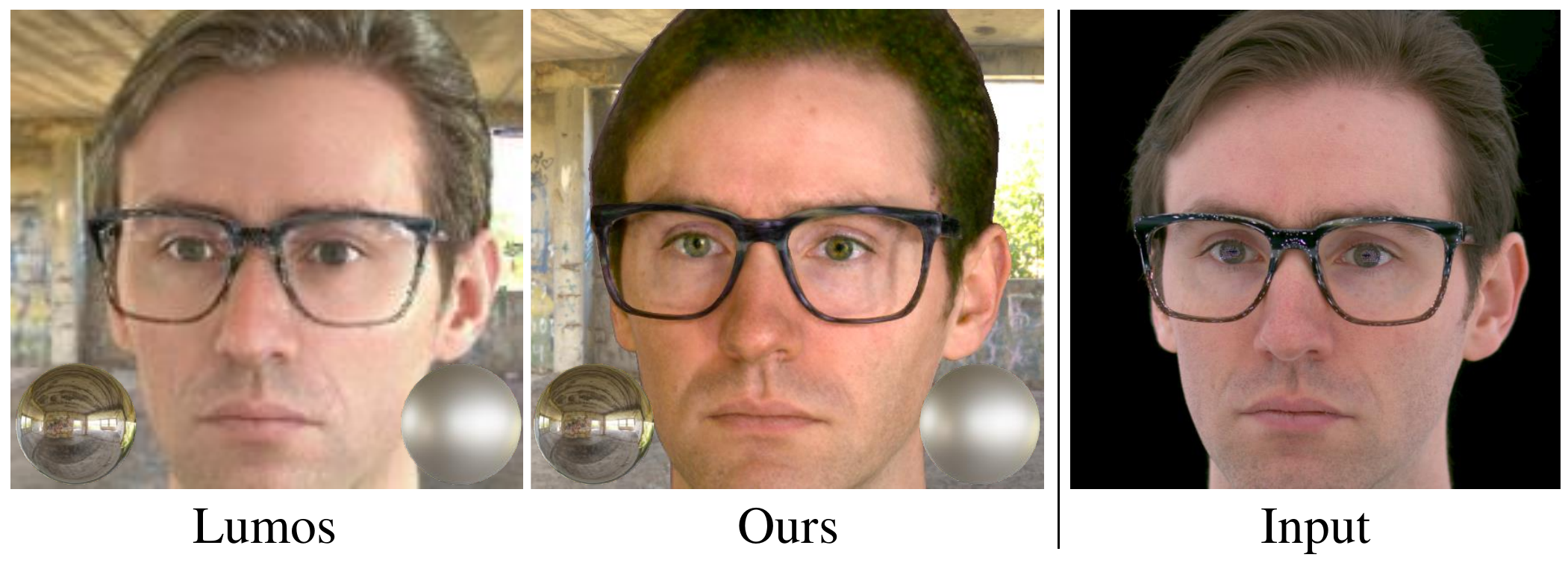}
    \vspace{-0.7cm}
    \caption{\textbf{Comparison with Lumos}~\cite{yeh2022learning}.
    Due to the 3D-aware lighting features, our method yields realistic shadows on the face. 
    }
    \label{fig:lumos}
    \vspace{-0.3cm}
\end{figure}

\subsection{Applications}
\label{sec:application}

\noindent
\textbf{Generative Eyeglasses.}
Our model is able to generate new eyeglasses via latent code modification (see supplementary video for more results). \cref{fig:change_shape} shows that our method supports replacing relightable materials while retaining shapes. 

\noindent
\textbf{Few-Shot Reconstruction.}
Our generative glasses model supports differentiable rendering, enabling few-shot reconstruction from a few-view images via inverse rendering. 
Notably, our non-relightable and relightable appearance models share the same latent codes.
Thus, as shown in \cref{fig:teaser}, the few-shot reconstruction using only fully-lit illumination can be rendered from novel illuminations. 

\noindent
\textbf{Lens Insertion.}
Since our model retains correspondences between primitives, inserting a lens in generated glasses is trivial by selecting control points for the lens contour on a single template. We further incorporate physically-accurate refraction and reflection based on prescription as shown in \cref{fig:teaser}. Please see \cref{sec:lensinsertion} for details of lens insertion implementation.

\begin{figure}
    \centering
    \includegraphics[width=0.48\textwidth]{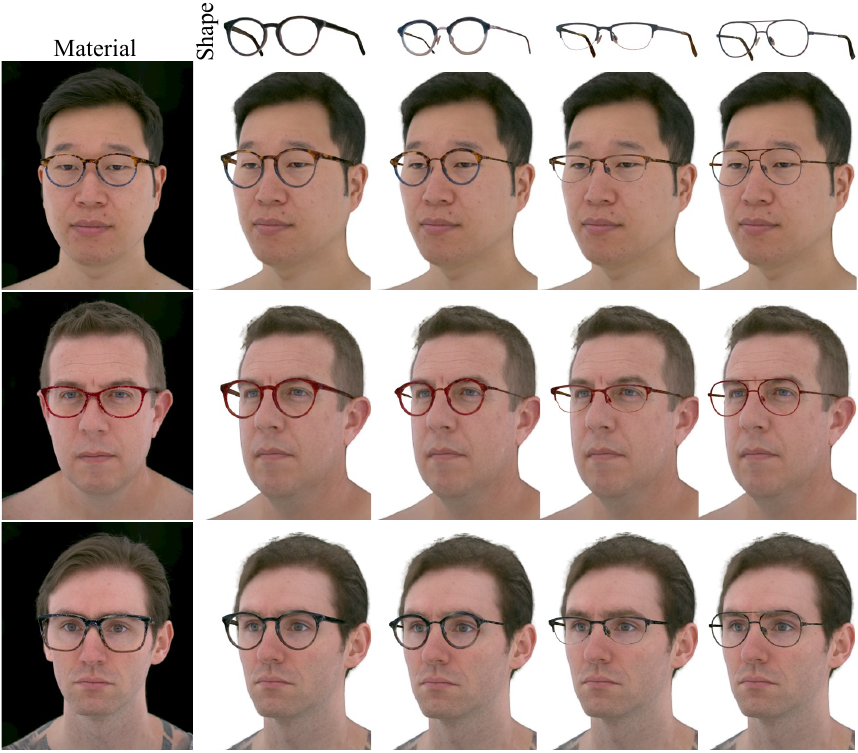}
    \caption{\textbf{Material swapping.} Our generative model supports changing materials and shape.
    }
    \label{fig:change_shape}
    \vspace{-0.3cm}
\end{figure}

\section{Conclusions}

We introduced MEGANE, a 3D morphable and relightable model of eyeglasses to create photorealistic compositions of eyeglasses on volumetric head avatars from any view point under novel illuminations.
Our experiments show that reproducing geometric and photometric interactions in the real world is now possible by leveraging neural rendering with a hybrid mesh-volumetric generative model.
By explicitly controlling the motion of primitives, our approach achieves, for the first time, the learning-based modeling of geometric interactions between glasses and faces.
We also examined the effectiveness of physics-inspired lighting features as inputs for neural relighting, and demonstrate that our approach enables relighting with a diverse set of materials that are both transmissive and reflective using a single generative model.
Lastly, we show that our generative model allows few-shot fitting to novel glasses, allowing relighting without additional OLAT data. 

Future work includes few-shot fitting to in-the-wild images by adopting a test-time finetuning as in \cite{cao2022authentic}, or physically accurate fitting of lenses via inverse rendering~\cite{li2020through,lyu2020differentiable}.

{\small
\bibliographystyle{ieee_fullname}
\bibliography{egbib}
}

\clearpage

\appendix
\section*{Supplementary Material}

\section{Data Acquisition}\label{sec:DataAcquisition}
In this section, we describe how we capture and process the \textit{Eyeglasses} dataset.

\subsection{Eyeglasses Dataset}
\begin{figure}
    \centering
    \includegraphics[width=0.48\textwidth]{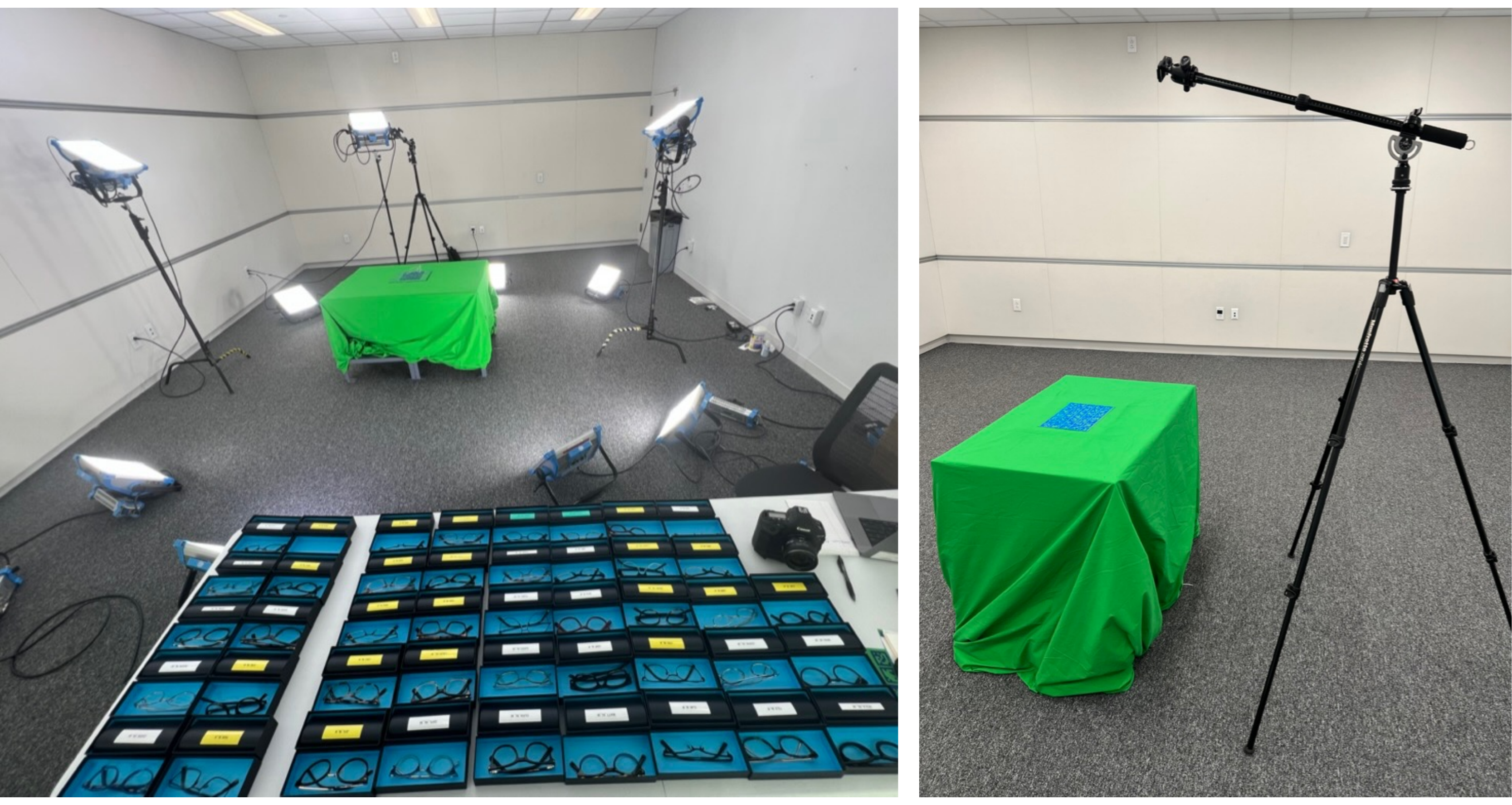}
    \caption{Our setup for capturing the \textit{Eyeglasses} dataset.}
    \label{fig:eyeglasses_dataset}
\end{figure}

\paragraph{Data Capture}
We capture the \textit{Eyeglasses} dataset consisting of 43 eyeglasses. The lenses from the eyeglasses were removed before capturing. We place the eyeglasses in a well lit indoor room. In addition, we place a AR-checker board with green background under the eyeglasses, as shown in \cref{fig:eyeglasses_dataset}. 
For each eyeglasses, we capture around 70 images from different view points with a hand-held DSLR camera. The camera intrinsics are calibrated in advance and fixed during the entire capture. We use the OpenCV detector~\cite{garrido2014automatic} and COLMAP~\cite{schoenberger2016sfm} for camera extrinsic and intrinsic calibration.

\paragraph{Mesh Extraction}
We employ NeuS~\cite{wang2021neus} to reconstruct the 3D mesh of each eyeglasses from the aforementioned multi-view capture. Specifically, we use the official NeuS implementation and its default hyper-parameters to train the network. NeuS was trained for 300k iterations with an NVIDIA V100 GPU, which takes around 8 hours. Once trained, a 3D mesh of the glasses can be extracted using marching cubes~\cite{Lorensen1987} with a grid resolution of $512$. We denote the meshes of the eyeglasses as
\begin{align}
	\gM_{i}   \in \R ^ {3\times M_i} ,\quad \gV_{i} \in \R ^ {3\times V_i},
\end{align}  
where $\gV_{i} $ are the vertices and $\gM_{i} $ are the faces of the $i$-th glasses.

\begin{figure}
    \centering
    \includegraphics[width=0.48\textwidth]{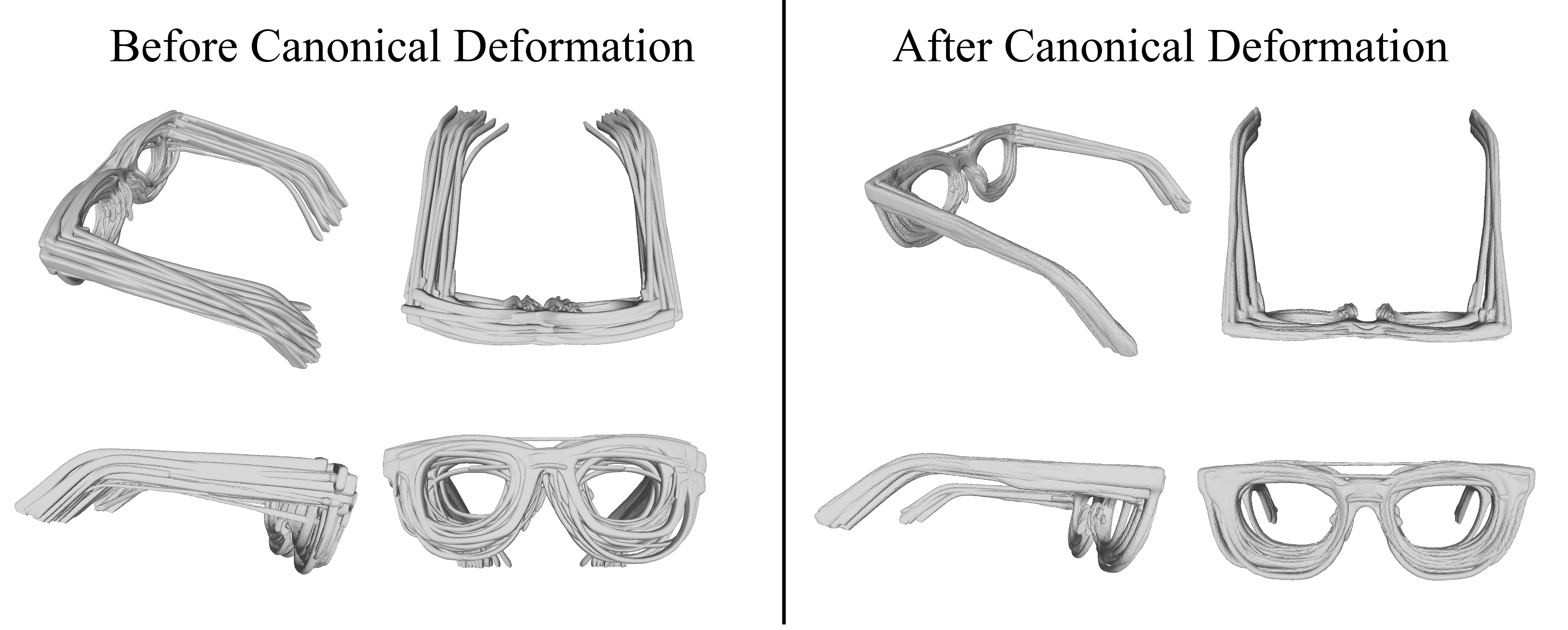}
    \caption{\textbf{Meshes of 43 eyeglasses.} The left side shows 43 eyeglasses extracted from NeuS, without spatial alignment. The right side shows the meshes in the canonical space.}
    \label{fig:raw_can_mesh}
\end{figure}

\paragraph{Mesh Canonicalization}
We deform these eyeglasses into a canonical space such that they are spatially aligned across different eyeglasses. 
We label a set of key points for each eyeglasses on 2D images, and then triangulate these 2D points to get the 3D key points $p_i \in \R^{3\times 20}$ on the eyeglasses mesh $\{  \gM_{i}   , \gV_{i} \}$.
We connect these key points to form a skeleton and apply Bounded Biharmonic Weights (BBW)~\cite{jacobson2011bounded} to deform the mesh into a canonical space using linear blend skinning (LBS).
Denote the linear blend skinning weights computed by BBW as $\mM_i \in \R^{20\times V_i}$ for eyeglasses $i$; we optimize the transformation of the  skeletons $\mT_i \in \R^{3\times 20}$ such that the L2 distance between the transformed key points and the average key points is minimized as follows:
\begin{align}
	\mT_i = \argmin_{\mT_i} || p^g_i  - \hat{p}||^2_2,
\end{align}
where the $p^g_i$ are the key points after applying the transformation; the transformed vertices of eyeglasses is given by $\gV^g_i = \mT_i \mM_i  $.
\cref{fig:raw_can_mesh} shows the effect of alignment. On the left are the extracted meshes of all 43 glasses, and the right is the deformed and transformed canonical meshes, where they are aligned based on the average key points.

\begin{figure}
    \centering
    \includegraphics[width=0.48\textwidth]{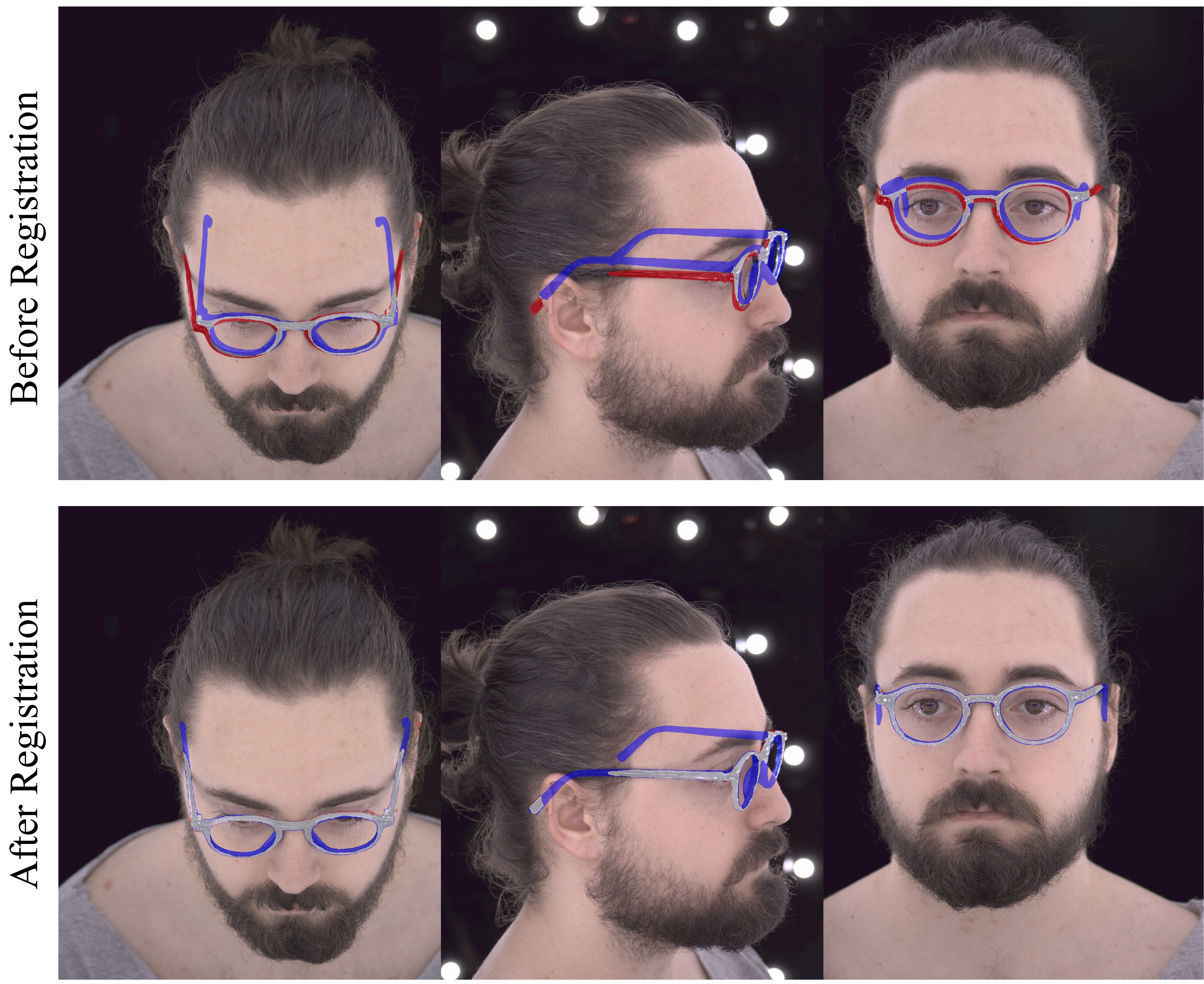}
    \caption{\textbf{Eyeglasses Registration on Face.} The top row shows the projection of canonical eyeglasses mesh on face with only rigid pose alignment. The bottom row shows eyeglasses mesh after non-rigid registration. The red color denote the detected segmentation of eyeglasses on face. The blue color in the figure denotes the projection of mesh. The mesh after person-dependent deformations align accurately with the observed images. }
    \label{fig:face_register}
\end{figure}

\paragraph{Glasses Registration on Face}
We now register the reconstructed meshes in the canonical space to fit the image data captured in the \textit{Faces with Eyeglasses} dataset. In this step we aim to model the person-dependent deformations of different eyeglasses on different people. 
For $j$-th subject wearing $i$-th eyeglasses, we compute the LBS weights of eyeglasses as $\mM_i$.
We choose one frame with neutral facial expression and regular eyeglasses position and fit the deformation of eyeglasses to this frame. 
We optimize a transformation and deformation matrix $\mA_{ij} $ such that the transformed/deformed mesh $\gV^g_{ij} = \mA_{ij}\mM_i$ has minimum key points loss and segmentation error:
\begin{align}
	\mA_{ij} = \argmin_{\mA_{ij}}\left(       || p^g_{ij}- p_j ||^2_2 + || \mI^g_{\text{seg}} - \mI_{\text{seg}} ||  \right),
\end{align}
where $p_j$ are the detected glasses key points on face wearing eyeglasses images; and $\mI_{\text{seg}}$ is the glasses segmentation on face wearing eyeglasses images; $\mI^g_{\text{seg}}$ is the rendered segmentation mask of the deformed eyeglasses mesh $\{ \gM_{i}   , \gV^g_{ij}   \}$. 

We use stochastic gradient descent with an Adam optimizer~\cite{kingma2014adam} to update the skeleton transformations with a learning rate of $10^{-3}$ for $1000$ iterations. The registration process takes around $20$ minutes for each eyeglasses. As shown in \cref{fig:face_register}, the deformed mesh after registration is aligned accurately with the observed images.

\begin{figure*}
    \centering
    \includegraphics[width=0.98\textwidth]{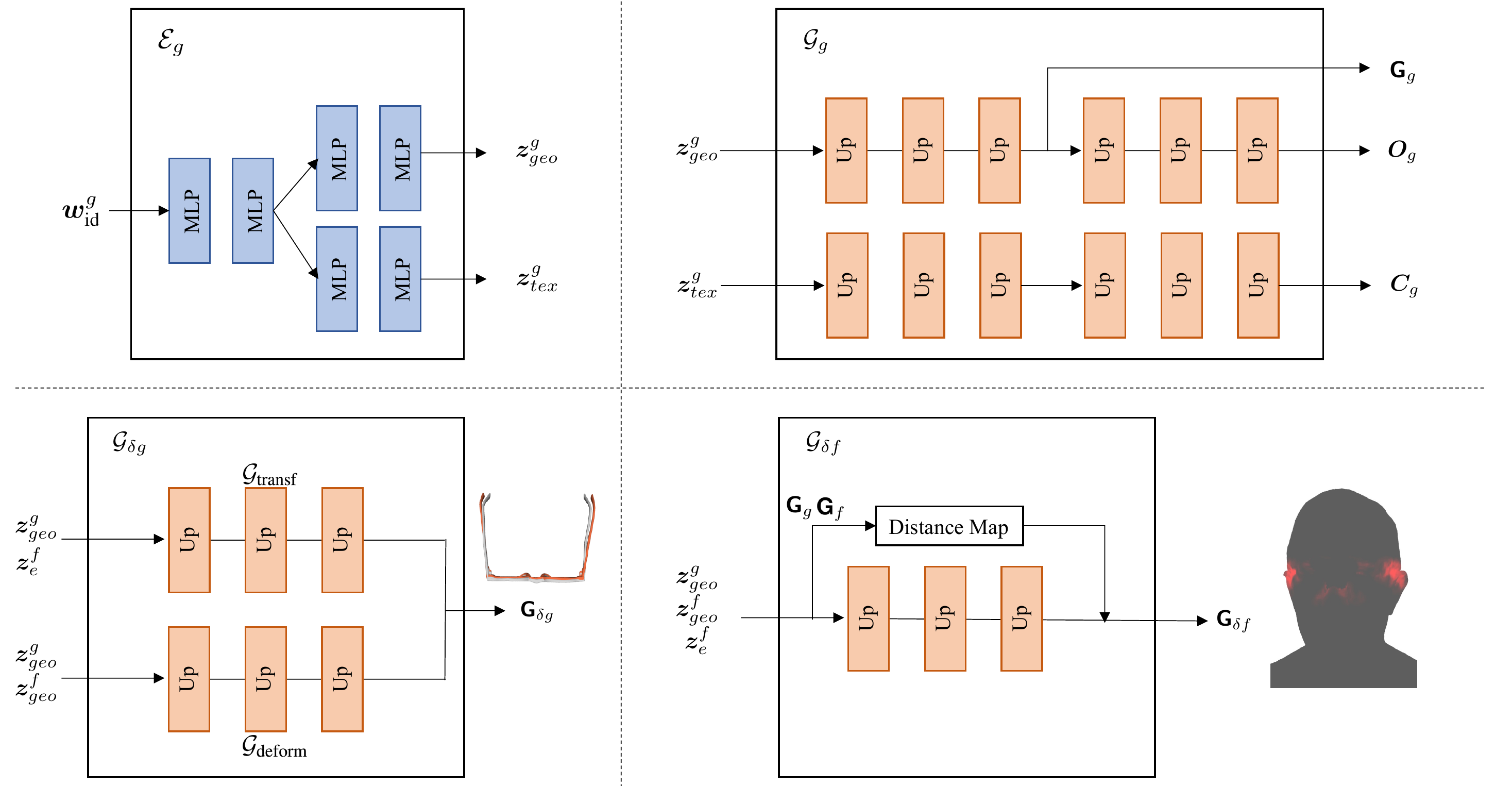}
    \caption{\textbf{Morphable Geometry Networks.} We illustrate the network architectures of $\gE_g, \gG_g, \gG_{\delta g}, \gG_{\delta f}$. The ``MLP'' in the figure denotes a linear layer followed by a leaky-ReLU with $0.2$ negative slope. The ``Up'' denotes an up-sampling layer consists of a transpose convolutional layer ($4 \times 4$ kernel, stride $2$), followed by a leaky-ReLU. The ``Distance Map'' computes the $l_2$ distance between each of the glasses primitives to its closest face primitives. 
    }
    \label{fig:networks1}
\end{figure*}

\section{Training and Losses}\label{sec:trainingandlosses}
In this section, we explain the loss function and training procedures in detail.

We denote all the ground truth camera images and associated processed assets for a frame $i$ as $\mI^i$,  which includes: the canonical mesh of the $i$-th eyeglasses $\{ \gM_{i} , \gV^g_{i} \}$; the deformed $i$-th mesh on $j$-th subject as  $ \{ \gM_{i} , \gV^g_{ij} \}  $; the mask of the canonical mesh $\mI^g_{\text{seg}}$; the mask of the deformed mesh $\mI^g_{ij\text{seg}}$;  the observed image $\mI$; glasses segmentation of observed image $\mI_{\text{seg}}$. 
We provide the exact formulation of each loss described in the main paper below as follows:
\begin{align}
	\gL_{\text{L1}} &=  || \mI' - \mI ||   ,  \\
	\gL_{\text{vgg}} &= \text{VGG} (\mI' , \mI ) ,\\
	\gL_{\text{gan}} &= \text{GAN} (\mI' , \mI )  ,
\end{align}
where $\mI' $ is the reconstructed image from volume rendering; and we follow the implementation of  $\text{VGG}(\cdot), \text{GAN}(\cdot) $ in~\cite{cao2022authentic}. In the notation below, we use prime $\mI'$ to denote the rendered results and notations without prime $\mI$ to denote the corresponding ground-truth.

\begin{align}
	\gL_{\text{c}} = \text{chamfer} (\gV^{g'}_{i} , \gV^g_{i} ) +  \text{chamfer} (\gV^{g'}_{ij} , \gV^g_{ij} )  , \\
	\gL_{\text{m}} = || \mI^{g'}_{\text{seg}} - \mI^g_{\text{seg}} || + || \mI^{g'}_{ij\text{seg}} - \mI^g_{ij\text{seg}} || , \\
	\gL_{\text{s}} = || \mI^{'}_{\text{seg}} - \mI_{\text{seg}}  || ,  
\end{align}
where $\text{chamfer}(\cdot)$ is the chamfer distance between two point clouds; $\gV^{g'}_{i} , \gV^{g'}_{ij} $ are the positions of eyeglasses primitives before and after person-dependent deformations
respectively; and  $ \mI^{g'}_{\text{seg}} ,  \mI^{g'}_{ij\text{seg}},  \mI^{'}_{\text{seg}}  $ are the rendered  eyeglasses mask, and segmentation of the corresponding eyeglasses deformations.

An $l_2$-regularization is also applied to the facial deformation terms
\begin{align}
	\gL_{\text{L2}} = || \delta \vs  ||^2_2 + || \delta \mR  ||^2_2 + || \delta \vt  ||^2_2.
\end{align}

The training of the network $\gA_{\text{spec}}$ relies on estimated normals $\vn$. For each eyeglasses mesh $\{\gM^g_{ij}, \gV^g_{ij}\}$, we extract its per-vertex surface normal and learn normals inside each primitive such that the predicted normals are coherent with the ones on the closest vertices.

During morphable geometry training, we first train the face model $\gG_f$ on face only dataset following~\cite{cao2022authentic}. We then jointly train other models on face wearing glasses dataset. Likewise, during relightable appearance training, we first train the face model $\gA_f$ on face only dataset; then we train other modules on the face wearing glasses dataset. We empirically found that the pretraining of the face modules is critical for stable training of the remaining modules including the interactions between faces and glasses.

\begin{figure*}
    \centering
    \includegraphics[width=0.98\textwidth]{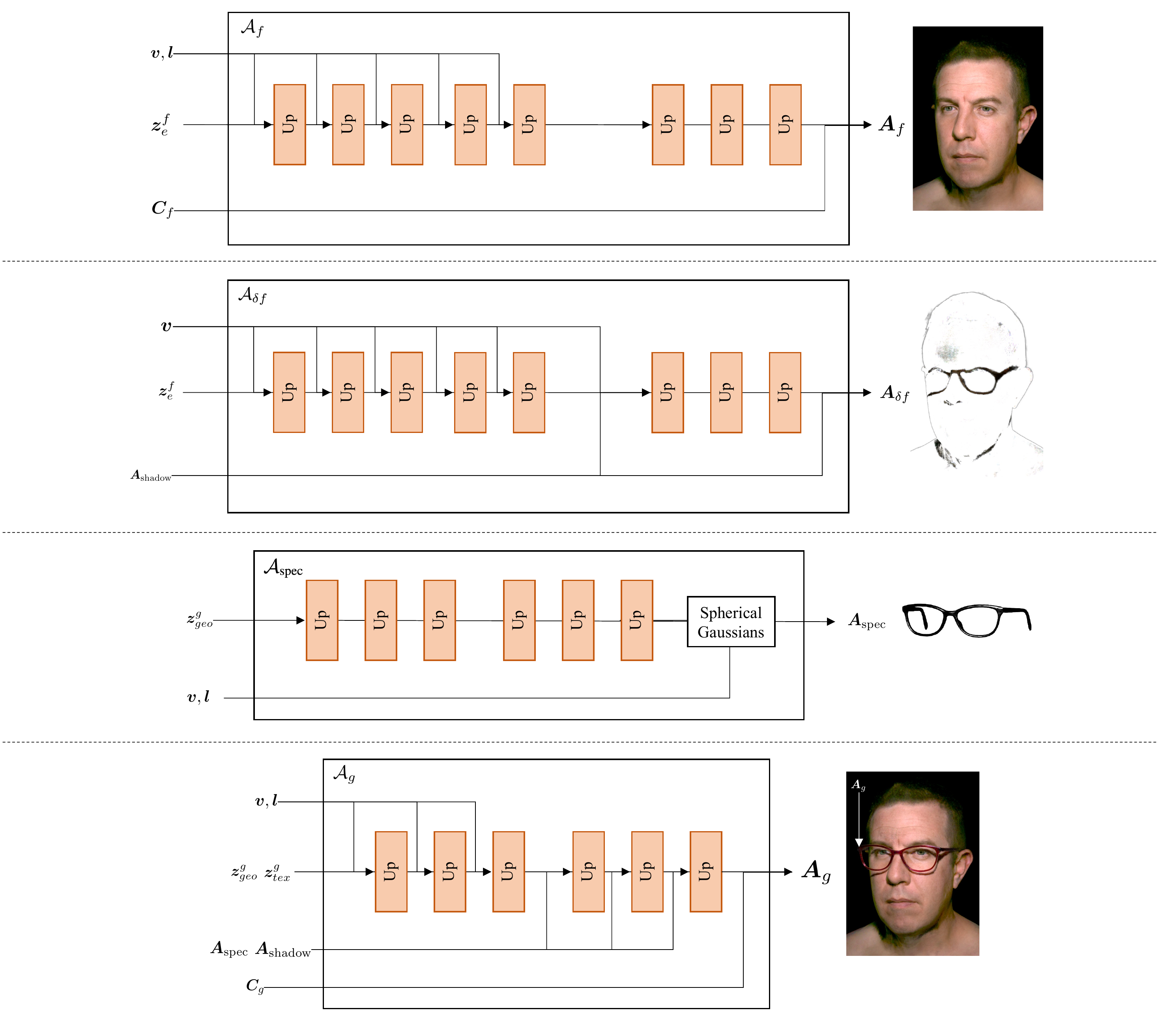}
    \caption{\textbf{Relightable Appearance Networks.} 
    We illustrate the network architectures of $\gA_f, \gA_{\delta f},  \gA_{\text{spec}}, \gA_{g}$. The ``Up'' denotes the same operation as in \cref{fig:networks1}. The ``Spherical Gaussians'' takes normal $\vn$, view direction $\vv$, and light direction $\vl$ as input, and computes three specular lobes via $s = \exp(r (\frac{\vl+\vv}{||\vl+\vv ||_2} \vn^T-1))$. In our experiments, we choose the following three roughness terms $r=\{64,128,1000\}$. 
    }
    \label{fig:networks2}
\end{figure*}

\section{Networks Architectures}
In this section, we provide the architectures of Morphable Geometry Networks and Relightable Appearance Networks in \cref{fig:networks1} and \cref{fig:networks2} separately.

\section{Lens Insertion}\label{sec:lensinsertion}
\begin{figure}
    \centering
    \includegraphics[width=0.48\textwidth]{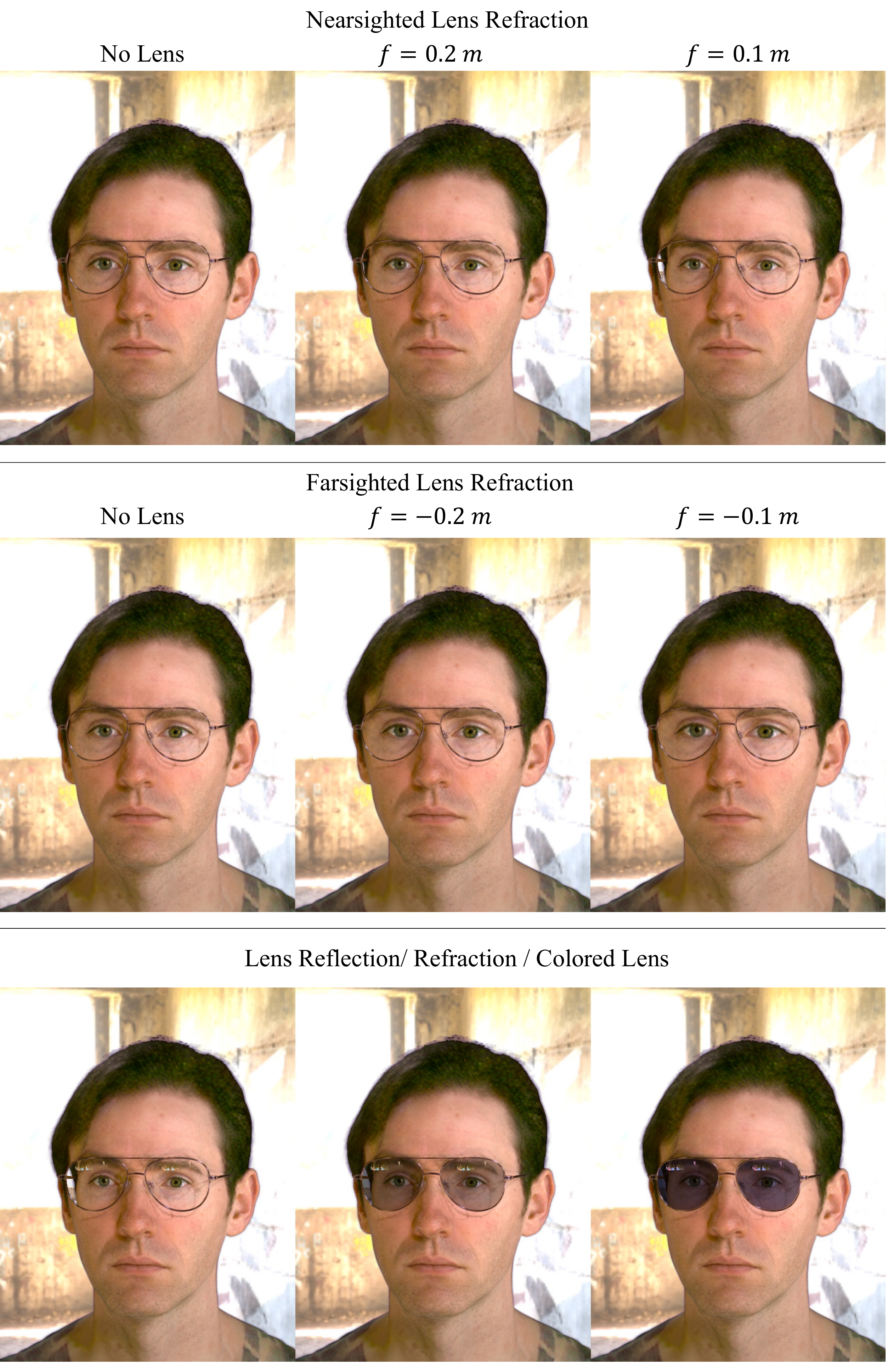}
    \caption{\textbf{Lens insertion.} Top row shows nearsighted lens refraction with different focal lengths. Second row shows farsighted lens refraction with different focal lengths. Bottom row shows the combination effects of the lens reflection, refraction and colored lens insertion. 
    }
    \label{fig:lens_insertion}
\end{figure}

We propose to model lenses as a postprocess by introducing an analytical model instead of jointly modeling them with eyeglasses frames from image observations.
The advantage of this analytical model of lens is that it yields plausible and photorealistic reflection and refraction for any prescription and doesn't required large dataset of lens for training. 
As shown in \cref{fig:lens_insertion}, we can even control the prescriptions of eyeglasses and intensity of the reflections. 

Without loss the generality, we focus on the left lens for explanation. A similar formulation is applied to the right lens.

\begin{figure}
    \centering
    \includegraphics[width=0.48\textwidth]{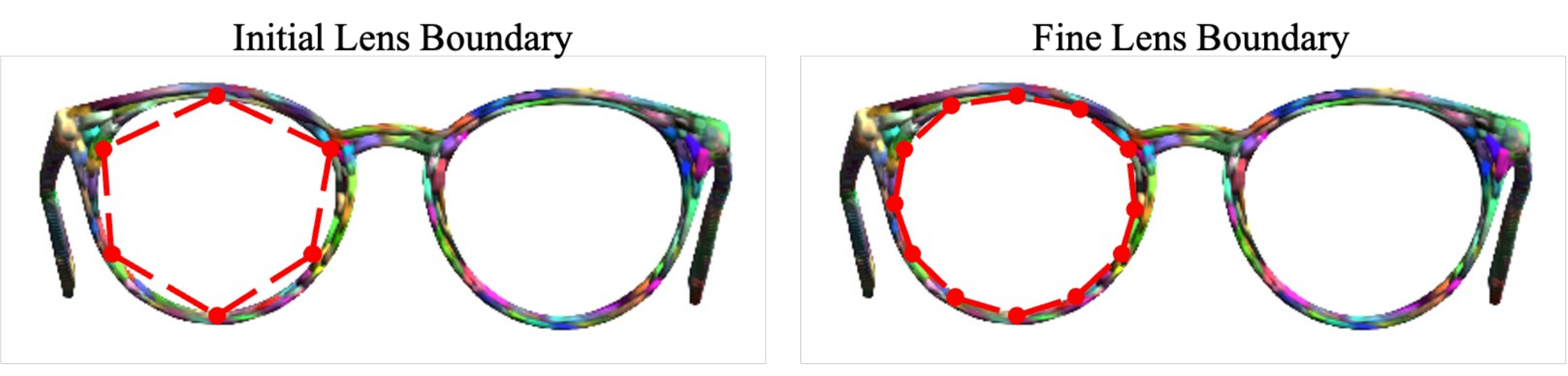}
    \caption{\textbf{Lens boundary estimation.} We demonstrate how the coarse boundary from glasses key points is refined to a finer lens boundary. Left is the six key points detected from images. Right is the fine lens boundary after one subdivision. In our experiments, we repeat the proposed subdivision three times.
    }
    \label{fig:lens_boundary}
\end{figure}

\begin{figure}
    \centering
    \includegraphics[width=0.48\textwidth]{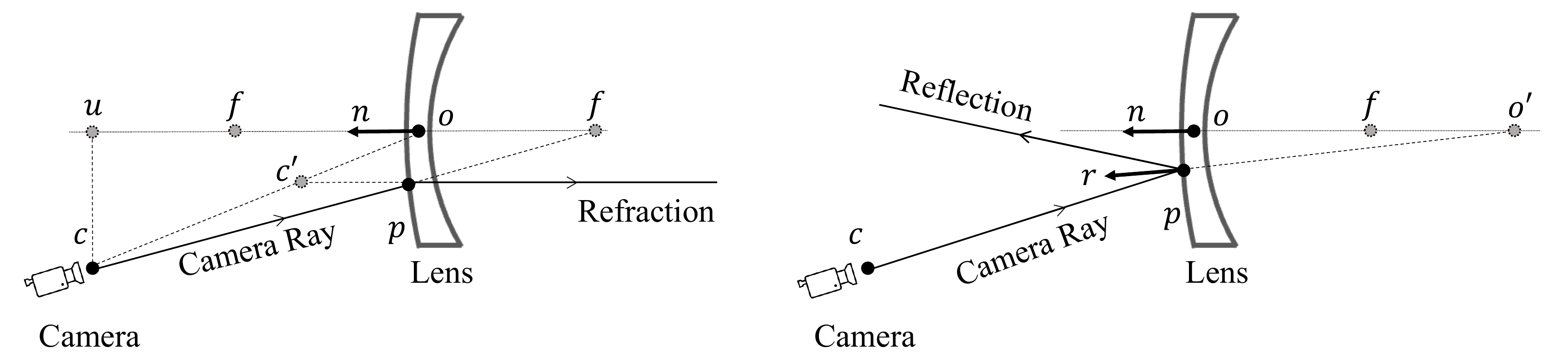}
    \caption{\textbf{Lens refraction and reflection.} The left and right shows how the ray refraction and reflection works in our lens insertion modeling.
    }
    \label{fig:refraction_draw}
\end{figure}

\smallskip
\noindent
\textbf{Lens boundary.}
We denote the detected key points of glasses in the image space. We first triangulate these key points to 3D. As shown in \cref{fig:lens_boundary}, the key points for left lens are not precise enough to draw lens. We therefore iteratively subdivide points and find the closest primitive positions. We apply the subdivision several times to obtain a fine lens boundary as shown in \cref{fig:refraction_draw}. 
With the estimated lens boundary, it is trivial to define a triangle mesh $m$ of lens by connecting the lens center with each boundary point.

\smallskip
\noindent
\textbf{Lens ray-marching for refraction and reflection.}
During ray marching, lens refraction and reflection only happens on those rays that are intersect with the lens mesh. Given the intersection point $\vp$ of the camera ray $\vd$ and lens mesh $m$, the distorted camera ray $\vd'$ is given by 
\begin{align}
	\vd' &=  \frac{\vp - \vc' }{||\vp - \vc' ||_2} , \\
	\vc' =       \frac{f (\vc - \vo)}{f + u}  &+ \vo, \quad  u = (\vc - \vo) \vn^T ,
\end{align}
where $\vc$ is the camera position; $f$ is the lens focal length, which can be derived from prescriptions; $\vn$ is the normal of lens mesh $m$;
$\vo$ is the optical center of lens, where we use the average of the lens boundary for approximation.

To compute the reflection direction, we approximate the lens as a sphere with radius $r$ as shown in \cref{fig:refraction_draw}. The reflection ray $\vd''$ can be computed as 
\begin{align}
	\vd'' = \vd - 2 (\vd \vr^T) \vr, \quad \vd = \frac{\vp-\vc}{||\vp - \vc||}, \\
    \vr = \frac{\vp-\vo'}{||\vp - \vo'||}, \quad  \vo' = \vo - r\vn.
\end{align}

When the camera ray is intersect with lens, we updated ray directions to a refraction ray and a reflective ray, and proceed the volume rendering, except when the ray does not intersect with any primitives, we query the sphere-mapped environment map.
Then the reflection and refraction are blended as 
\begin{align}
	I = \alpha I_{\text{refra}} + \beta I_{\text{refle}},
\end{align}
where $ \alpha, \beta$ is the ratio of refraction and reflection respectively.

\section{Few-Shot Reconstruction}
Our method is fully differentiable. Hence, once trained, we can use only a few images to reconstruct the geometry and material of novel glasses unseen during training.
Specifically, we optimize the geometry latent code and appearance latent code via photometric loss:
\begin{align}
	z^g_{\text{geo}}, z^g_{\text{tex}} = \argmin_{z^g_{\text{geo}}, z^g_{\text{tex}}} \sum || \mI' - \mI || ,
\end{align}
where $\mI$ is the observed images; and $\mI'$ is the rendered images using the latent code $z^g_{\text{geo}}, z^g_{\text{tex}}$.

\section{Implementation details of comparisons}
In this section, we explain implementation details of our comparisons including the modifications we make to GeLaTO~\cite{martin2020gelato} and GIRAFFE~\cite{niemeyer2021giraffe} to support our own dataset. 

\subsection{Implementation of GeLaTO~\cite{martin2020gelato}}
GeLaTO is not open sourced, and their training dataset is also not released. Therefore, we implement their method following their paper and train using our \textit{Faces with Eyeglasses} dataset. We use the detected segmentation of glasses as a ground-truth foreground mask. To align the three billboards proposed in their work, we use the detected 3D key points as in our training pipeline for fair comparison.

\subsection{Implementation of GIRAFFE~\cite{niemeyer2021giraffe} }
GIRAFFE proposed a compositional neural radiance field that supports adding and changing objects in a scene. However, the official implementation only supports objects within the same category. For a fair comparison, we adapt their method to support adding generative objects in multiple categories. 

Specifically, their official implementation supports only two models: a background model and a object model. We extend this to support three different models for a background, faces, and glasses. To further facilitate the decomposition of faces and glasses, we combine \textit{Faces only} and \textit{Faces with Eyeglasses} datasets for training. Without this, we observe that GIRAFFE learns to model faces and glasses in a single model as they are always co-located.

\subsection{Implementation of Envmap Relighting }
Following Bi~\etal~\cite{bi2021deep}, we represent environmental lights as a set of distant lights, and compute shadow features for each light source. Due to the linearity of light transport, we can synthesize faces and eyeglasses under arbitrary environmental lights by linearly blending contributions of each light. Note that the global intensity and color balance may not be consistent between ours and Lumos because Lumos does not release their tone mapping function or global intensity scale.

\section{Limitations}
While our model successfully models the deformation residuals on glasses conditioned by face identity and expressions, the initial position of glasses and subtle motions caused by facial expression changes are entangled. Future work could address this limitation by incorporating more fine-grained data capture and loss functions to facilitate disentanglement. Another limitation is that our current framework infers relighting results under a single point light. On one hand, due to the linearity of light transport~\cite{debevec2000acquiring}, we can synthesize physically plausible relighting under natural illuminations by weighted sum of multiple point light sources. On the other hand, running the relighting network for each point light source is too expensive for real-time use. As demonstrated for face relighting \cite{bi2021deep}, distilling our point-light based model to an efficient student model should be possible for real-time use cases.

\end{document}

%% file: math_commands.tex
\usepackage{amsmath,amsfonts,bm}

\def\eqref#1{equation~\ref{#1}}

\def\1{\bm{1}}

\def\vc{{\bm{c}}}
\def\vd{{\bm{d}}}

\def\vl{{\bm{l}}}

\def\vn{{\bm{n}}}
\def\vo{{\bm{o}}}
\def\vp{{\bm{p}}}

\def\vr{{\bm{r}}}
\def\vs{{\bm{s}}}
\def\vt{{\bm{t}}}

\def\vv{{\bm{v}}}
\def\vw{{\bm{w}}}

\def\vz{{\bm{z}}}

\def\mA{{\bm{A}}}

\def\mC{{\bm{C}}}

\def\mI{{\bm{I}}}

\def\mM{{\bm{M}}}

\def\mO{{\bm{O}}}

\def\mR{{\bm{R}}}

\def\mT{{\bm{T}}}

\DeclareMathAlphabet{\mathsfit}{\encodingdefault}{\sfdefault}{m}{sl}
\SetMathAlphabet{\mathsfit}{bold}{\encodingdefault}{\sfdefault}{bx}{n}
\newcommand{\tens}[1]{\bm{\mathsfit{#1}}}

\def\tG{{\tens{G}}}

\def\gA{{\mathcal{A}}}

\def\gE{{\mathcal{E}}}

\def\gG{{\mathcal{G}}}

\def\gL{{\mathcal{L}}}
\def\gM{{\mathcal{M}}}

\def\gV{{\mathcal{V}}}

\def\sR{{\mathbb{R}}}

\newcommand{\R}{\mathbb{R}}

\DeclareMathOperator*{\argmin}{arg\,min}